\pdfoutput=1

\documentclass[11pt]{article}

\usepackage[preprint]{acl}

\usepackage{times}
\usepackage{latexsym}

\usepackage[T1]{fontenc}

\usepackage[utf8]{inputenc}

\usepackage{microtype}

\usepackage{inconsolata}

\usepackage{graphicx}
\usepackage{CJKutf8}
\usepackage{tabularx}
\usepackage{booktabs}
\usepackage{array}
\usepackage{enumitem}
\usepackage{multirow}
\usepackage{colortbl}
\usepackage{pifont}
\usepackage{float}
\usepackage{marvosym}
\usepackage[most]{tcolorbox}

\newcommand{\cmark}{\ding{51}}
\newcommand{\xmark}{\ding{55}}
\newcommand{\dashmark}{--}
\newcommand{\mQwenThreeTwoThreeFive}{\href{https://huggingface.co/Qwen/Qwen3-235B-A22B}{Qwen3-235B-A22B}}
\newcommand{\mKimiKTwoFive}{\href{https://huggingface.co/moonshotai/Kimi-K2.5}{Kimi-K2.5}}
\newcommand{\mQwenThreeThirty}{\href{https://huggingface.co/Qwen/Qwen3-30B-A3B-Thinking-2507}{Qwen3-30B-A3B}}
\newcommand{\mQwenThreeFiveThreeNineSeven}{\href{https://huggingface.co/Qwen/Qwen3.5-397B-A17B}{Qwen3.5-397B-A17B}}
\newcommand{\mDeepSeekROne}{\href{https://huggingface.co/deepseek-ai/DeepSeek-R1}{DeepSeek-R1}}
\newcommand{\mQwenNemotron}{\href{https://huggingface.co/nvidia/Qwen3-Nemotron-8B-BRRM}{Qwen3-Nemotron-8B-BRRM}}
\newcommand{\mSkyworkQwenThree}{\href{https://huggingface.co/Skywork/Skywork-Reward-V2-Qwen3-8B}{Skywork-Reward-V2-Qwen3-8B}}
\newcommand{\mDeepSeekVThreeTwo}{\href{https://huggingface.co/deepseek-ai/DeepSeek-V3.2}{DeepSeek-v3.2}}
\newcommand{\mQwenThreeFourteen}{\href{https://huggingface.co/Qwen/Qwen3-14B}{Qwen3-14B}}
\newcommand{\mQwenThreeEight}{\href{https://huggingface.co/Qwen/Qwen3-8B}{Qwen3-8B}}
\newcommand{\mQwenThreeFiveTwentySeven}{\href{https://huggingface.co/Qwen/Qwen3.5-27B}{Qwen3.5-27B}}
\newcommand{\mQwenThreeFour}{\href{https://huggingface.co/Qwen/Qwen3.5-4B}{Qwen3-4B}}
\newcommand{\mLegalOne}{\href{https://huggingface.co/CSHaitao/LegalOne-8B}{LegalOne-8B}}
\newcommand{\mKimiKTwo}{\href{https://huggingface.co/moonshotai/Kimi-K2-Instruct-0905}{Kimi-K2}}
\newcommand{\mGemmaThreeTwelve}{\href{https://huggingface.co/google/gemma-3-12b-it}{Gemma-3-12B-IT}}
\newcommand{\mSkyworkLlamaThreeTwo}{\href{https://huggingface.co/Skywork/Skywork-Reward-V2-Llama-3.2-3B}{Skywork-Reward-V2-Llama-3.2-3B}}
\newcommand{\mSkyworkLlamaThreeOne}{\href{https://huggingface.co/Skywork/Skywork-Reward-V2-Llama-3.1-8B}{Skywork-Reward-V2-Llama-3.1-8B}}
\newcommand{\mRRMSeven}{\href{https://huggingface.co/Reward-Reasoning/RRM-7B}{RRM-7B}}
\newcommand{\mLlamaRM}{\href{https://huggingface.co/allenai/Llama-3.1-8B-Instruct-RM-RB2}{Llama-3.1-8B-Instruct-RM-RB2}}
\newcommand{\mRMRQwen}{\href{https://huggingface.co/gaotang/RM-R1-Qwen2.5-Instruct-7B}{RM-R1-Qwen2.5-Instruct-7B}}
\newcommand{\mQwenThreeOneSeven}{\href{https://huggingface.co/Qwen/Qwen3-1.7B}{Qwen3-1.7B}}
\newcommand{\mLlamaThreeOne}{\href{https://huggingface.co/meta-llama/Llama-3.1-8B-Instruct}{Llama-3.1-8B-Instruct}}
\newcommand{\mSkyworkLlamaVZeroTwo}{\href{https://huggingface.co/Skywork/Skywork-Reward-Llama-3.1-8B-v0.2}{Skywork-Reward-Llama-3.1-8B-v0.2}}
\newcommand{\mQwenTwoFiveSeven}{\href{https://huggingface.co/Qwen/Qwen2.5-7B-Instruct}{Qwen2.5-7B-Instruct}}
\newcommand{\mInternLMReward}{\href{https://huggingface.co/internlm/internlm2-1_8b-reward}{InternLM2-1.8B-Reward}}
\newcommand{\mMiniCPMFour}{\href{https://huggingface.co/openbmb/MiniCPM4-8B}{MiniCPM4-8B}}
\newcommand{\mLlamaThreeTwo}{\href{https://huggingface.co/meta-llama/Llama-3.2-3B-Instruct}{Llama-3.2-3B-Instruct}}
\newcommand{\mQwenThreeZeroSix}{\href{https://huggingface.co/Qwen/Qwen3-0.6B}{Qwen3-0.6B}}

\newtcolorbox[auto counter]{promptbox}[1][]{
  breakable,
  enhanced,
  colback=gray!5,
  colframe=gray!50,
  coltitle=black,
  colbacktitle=gray!25,
  fonttitle=\small\bfseries\scshape,
  toptitle=3pt,
  bottomtitle=3pt,
  fontupper=\small,
  sharp corners,
  boxrule=0.5pt,
  left=5pt, right=5pt, top=5pt, bottom=5pt,
  title={Prompt~\thetcbcounter},
  #1
}

\title{\textsc{LegalRewardBench}: Benchmarking Reward Models for Legal Preference Evaluation}

\author{
    Hui Huang\textsuperscript{1}$^{*}$,
    Jie Zhou\textsuperscript{2}$^{*}$, 
    Hongli Zhou\textsuperscript{1},
    Qundong Shi\textsuperscript{2},
    Shuyuan Zheng\textsuperscript{3}, \\
    \textbf{Yini Su\textsuperscript{2}},
    \textbf{Shaochun Wang\textsuperscript{2}},
    \textbf{Biyuan Lin\textsuperscript{2}},
    \textbf{Muyun Yang\textsuperscript{1}\textsuperscript{\Letter}},
    \textbf{Tiejun Zhao\textsuperscript{1}} \\[5pt]    
    \textsuperscript{1}Harbin Institute of Technology, \textsuperscript{2}ModelBest Inc., \textsuperscript{3}The University of Osaka \\[3pt]
    \texttt{\{zhoujie, shiqundong, linbiyuan\}@modelbest.cn},\\
    \texttt{huanghui@stu.hit.edu.cn, yangmuyun@hit.edu.cn}
}

\title{\textsc{LegalRewardBench}: Benchmarking Reward Models for Legal Preference Evaluation}

\begin{document}
\maketitle

\let\oldthefootnote\thefootnote

\let\thefootnote\relax\footnotetext{$*$ Equal contribution.}
\let\thefootnote\oldthefootnote

\renewcommand{\thefootnote}{\fnsymbol{footnote}} 
\renewcommand{\thefootnote}{\arabic{footnote}} %

\begin{abstract}
Reward models are central to preference alignment for legal tasks, yet existing reward model benchmarks are still primarily centered on general-domain preferences. In this work, we introduce \textsc{LegalRewardBench}, the first human-verified benchmark for evaluating reward models in the legal domain. The benchmark organizes legal preference evaluation\footnote{In this paper, ``legal preference'' refers to human preferences over responses to legal queries, rather than the traditional legal inclinations in legal theory or litigation.} into five capability levels: \texttt{knowledge}, \texttt{understanding}, \texttt{analysis}, \texttt{decision}, and \texttt{writing}, and preference examples are constructed through rubric-guided automatic judging combined with human verification. We evaluate three families of judges, covering both general and fine-tuned models, and generative and discriminative architectures. Experimental results show that strong general-purpose generative models are currently the most robust legal reward models, while existing fine-tuned reward models transfer less reliably to legal preference evaluation. We also conduct detailed comparisons of input organizations and training strategies.

\end{abstract}

\section{Introduction}

\begin{figure}[t]
\centering
\includegraphics[width=\linewidth]{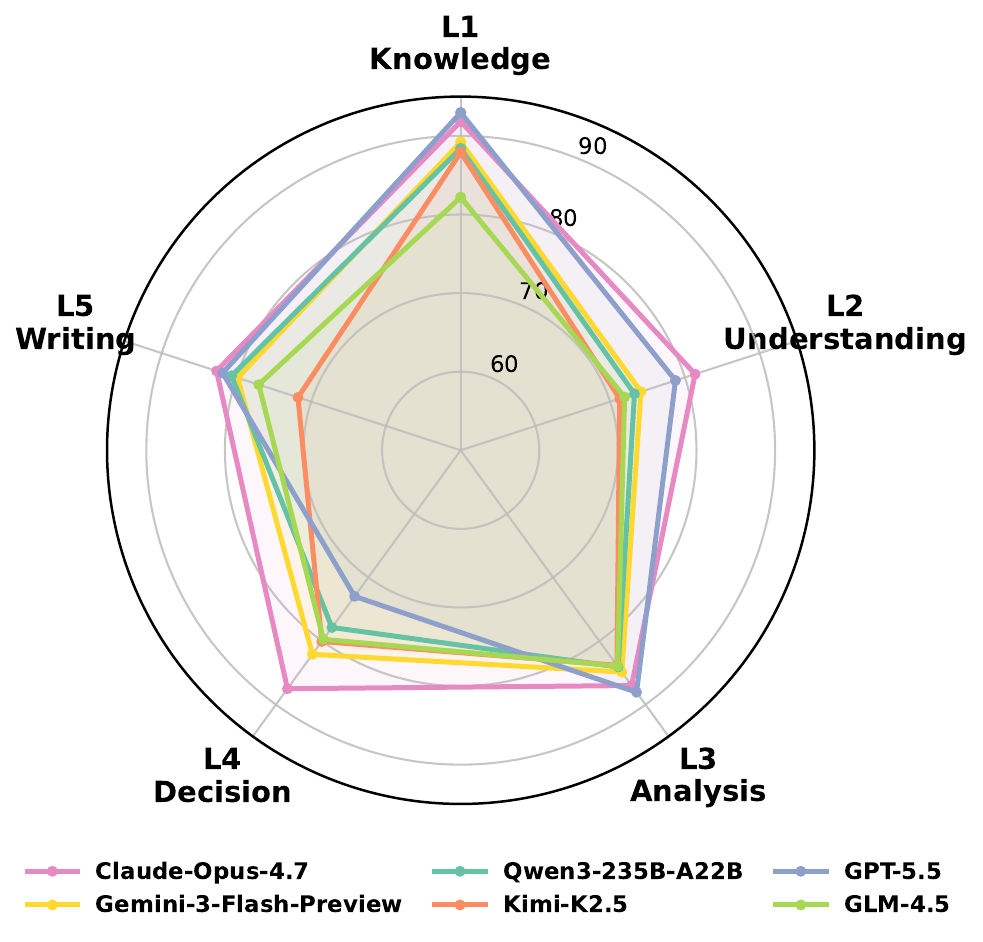}
\caption{Performance of best generative RMs across the five \textsc{LegalRewardBench} capability levels.}
\label{fig:model_level_radar}
\end{figure}




The rapid advancement of large language models (LLMs) has spurred growing interest in applying them to the legal domain~\cite{cui2023chatlaw, colombo2024saullm}. Existing work has explored a broad spectrum of tasks, including legal judgment prediction~\cite{zhong2020iteratively}, legal fact prediction~\cite{liu2025legal}, and court opinion generation~\cite{dahl2024large}. Despite significant progress, benchmarking results consistently reveal that even state-of-the-art LLMs fall short of satisfactory performance~\cite{fei2024lawbench, guha2023legalbench}, underscoring the need for principled methods to improve LLM capabilities in the legal domain.

Reinforcement learning from human feedback (RLHF) has emerged as a standard paradigm for LLM post-training~\cite{ouyang2022training, bai2022training}. At the heart of RLHF lies the reward model (RM), which serves as a learned proxy for human preferences and guides the policy model toward preferred behaviors~\cite{zhong2025comprehensive,ijcai2026p681}. 
Since legal reasoning depends on the theoretical and experiential preferences of legal experts, RLHF holds strong potential for strengthening LLM performance on legal tasks.
However, benchmarks for evaluating how well RMs capture legal preferences remain underexplored, leaving the field without a reliable means to measure advances in this direction.

To address this gap, we introduce \textsc{LegalRewardBench}, the first human-verified benchmark specifically designed to evaluate RMs on legal preference assessment. \textsc{LegalRewardBench} covers five core dimensions of legal preference---\textit{knowledge}, \textit{understanding}, \textit{analysis}, \textit{decision}, and \textit{writing}. We leverage a rubric-guided preference annotation strategy, combining both automatic scoring and human verification to ensure annotation precision.

Based on \textsc{LegalRewardBench}, we evaluate three families of RMs: general LLMs, fine-tuned generative RMs, and fine-tuned discriminative RMs. The results show that \textbf{strong general-purpose LLMs are currently the most robust choice for legal reward modeling, while many fine-tuned RMs transfer only weakly to legal preference evaluation}. Ablation experiments further show that reference answers substantially improve legal reward modeling. We further compare training methods, showing that legal RM training brings limited and inconsistent gains. These findings collectively demonstrate that legal reward modeling should be evaluated as a domain-specific problem rather than assumed to transfer from general preference benchmarks.

Our contributions are threefold:
\begin{itemize}[leftmargin=4mm, itemsep=1mm, parsep=0pt]
    \item We introduce \textsc{LegalRewardBench}, the first human-verified benchmark for legal reward modeling, via a rubric-guided preference annotation strategy;
    \item We conduct a systematic evaluation of existing reward-model families, showing that general models mostly outperform fine-tuned RMs;
    \item We conduct detailed ablation studies on inference and training strategies, offering guidance for reward modeling in the legal domain.
\end{itemize}

\section{Data Construction}


The construction pipeline of \textsc{LegalRewardBench} has three stages: 1) query collection, 2) response generation, and 3) preference annotation.


\subsection{Query Collection}

\textsc{LegalRewardBench} is organized around five legal capability levels inspired by Bloom's taxonomy \cite{krathwohl2002revision} and by prior legal evaluation benchmarks. As shown in Table \ref{tab:data_sources}, we prioritize source tasks with relatively clear evaluation criteria\footnote{The detailed query task distribution and data source information are presented in Appendix \ref{app:task_description}.}, categorized into five levels:

\begin{table}[t]
\centering
\small
\begin{tabularx}{\columnwidth}{>{\raggedright\arraybackslash}p{0.29\columnwidth}X}
\toprule
\textbf{Level} & \textbf{Representative Tasks} \\
\midrule
L1\newline \texttt{Knowledge} & Statute Recitation; Legal Rule Recall; Applicable Law Identification; Legal Knowledge Error Detection... \\
\midrule
L2\newline \texttt{Understanding} & Case Event Timeline Extraction; Legal Element Extraction; Long-Form Legal Document Understanding... \\
\midrule
L3\newline \texttt{Analysis} & Fact-Finding Analysis; Disputed Issue Identification; Evidence Analysis... \\
\midrule
L4\newline \texttt{Decision} & Sentence Prediction; Judgment Outcome Prediction; Legal Task Planning... \\
\midrule
L5\newline \texttt{Writing} & Procuratorial Recommendation Drafting; Defense Pleading Drafting; Indictment Drafting... \\
\bottomrule
\end{tabularx}
\caption{Representative tasks for the five capability levels in \textsc{LegalRewardBench}.}
\label{tab:data_sources}
\end{table}

\begin{enumerate}[leftmargin=4mm, itemsep=1mm, parsep=0pt]
    \item \textbf{\texttt{Knowledge:}} evaluates whether a model can recall, identify, and apply determinate legal knowledge in legally grounded contexts.
    \item \textbf{\texttt{Understanding:}} evaluates whether a model can read, preserve, and reorganize legal information from structured or unstructured legal texts.
    \item \textbf{\texttt{Analysis:}} evaluates whether a model can decompose legal problems and reason over facts, evidence, arguments, and legal categories.
    \item \textbf{\texttt{Decision:}} evaluates whether a model can make or simulate decision-oriented legal judgments under explicit task constraints.
    \item \textbf{\texttt{Writing:}} evaluates whether a model can produce structured legal documents under professional legal-writing conventions.
\end{enumerate}

This hierarchy enables fine-grained analysis of RMs across task types. It also makes the final benchmark more diagnostic: a model can be strong at identifying surface-level legal facts while still failing on long-form writing.

The collected queries mainly come from three sources: 1) in-house legal task data; 2) existing legal benchmarks; 3) previous legal exams. Each query is required to have a reference answer.

\subsection{Response Generation}

\begin{table*}[t]
\centering
\small
\resizebox{\textwidth}{!}{
\begin{tabular}{lcccccccc}
\toprule
\multirow{2}{*}{\textbf{Level}} & \multirow{2}{*}{\textbf{Samples}} & \multicolumn{3}{c}{\textbf{Average Length}} & \multicolumn{2}{c}{\textbf{Human-Human Agree.}} & \multicolumn{2}{c}{\textbf{Human-Model Agree.}} \\
\cmidrule(lr){3-5} \cmidrule(lr){6-7} \cmidrule(lr){8-9}
 & & \textbf{Query} & \textbf{Chosen} & \textbf{Rejected} & \textbf{w/ Tie} & \textbf{w/o Tie} & \textbf{w/ Tie} & \textbf{w/o Tie} \\
\midrule
L1 \texttt{Knowledge} & 306 & 234.5 & 449.9 & 435.9 & 76.19 & 76.19 & 81.69 & 86.31 \\
L2 \texttt{Understanding} & 228 & 1497.9 & 670.5 & 566.9 & 70.00 & 77.78 & 73.54 & 78.68 \\
L3 \texttt{Analysis} & 377 & 1292.3 & 661.7 & 446.2 & 64.71 & 64.71 & 77.71 & 81.15 \\
L4 \texttt{Decision} & 296 & 566.9 & 648.4 & 345.5 & 76.19 & 88.89 & 82.35 & 88.05 \\
L5 \texttt{Writing} & 124 & 6355.7 & 995.0 & 874.7 & 96.15 & 96.15 & 90.00 & 91.67 \\
\midrule
Overall & 1331 & 1394.7 & 642.6 & 482.0 & 79.20 & 83.90 & 79.93 & 86.12 \\
\bottomrule
\end{tabular}}
\caption{Data statistics and annotation agreement by capability level on \textsc{LegalRewardBench}.}
\label{tab:data_stats}
\end{table*}

To expose reward models to realistic response distributions, we ask a diverse set of models to generate responses for each legal task, and then construct preference pairs from the generated candidates. The candidate generation stage covers roughly 30 models, including open and closed models from more than 10 model families, with model sizes ranging from 3B to 1T parameters\footnote{Detailed information and inference settings about candidate models are provided in Appendices~\ref{app:model_catalog} and~\ref{app:inference_settings}. Note that we use reasoning-augmented inference when a model is a hybrid reasoning model or has a reasoning-augmented variant.}. This design increases response diversity, exposes a wider range of legal error types, and avoids overly homogeneous preference pairs.

Many legal queries originally ask only for a final answer, but in reward modeling the reasoning process often reveals errors that are invisible in the final answer alone. We therefore rewrite queries of all levels (except for \texttt{writing}) to ask the model to produce responses in a constrained format with analysis followed by the final answer. This preserves final-answer checkability while making legal reasoning available to preference annotation.


\subsection{Preference Annotation}
\label{sec:annotation}

The main challenge in legal preference annotation is to make judgments stable, interpretable, and auditable. We therefore introduce a \textbf{Rubric-guided Annotation} scheme\footnote{Detailed automatic-scoring and human-verification guidelines are provided in Appendix~\ref{app:annotation_guidelines}.}, with the following steps:

\begin{enumerate}[leftmargin=4mm, itemsep=1mm, parsep=0pt]
    \item \textbf{Rubric Construction}: we first ask legal-domain annotators to construct rubrics based on the reference answer, task requirements, and legal background knowledge. The rubric clearly defines the essential criteria required to constitute a comprehensive and legally sound response.
    \item \textbf{Automatic Scoring:} for verifiable tasks, such as sentence prediction, we design rules to score the candidate responses. For non-verifiable queries, we use a proprietary model to score candidate responses according to the rubrics. 
    \item \textbf{Pair Selection:} based on the automatic scoring, we construct pairwise examples whose score gap is sufficiently clear while avoiding overly trivial pairs, using the criterion that the chosen score should exceed the rejected score by at least 10\% and should itself rank in the top 50\%.
    \item \textbf{Human Verification:} each pair is assigned to an annotator to evaluate the queries and candidate responses, including their reasoning processes against the predefined rubrics to validate the correctness of the preference. We retain only those samples where the human verification aligns with the automatic preference.
\end{enumerate}

The data statistics are presented in the next section. Compared with purely subjective preference labels, this rubric-based design helps constrain the evaluation around legal criteria.

\subsection{Data Statistics}

As shown in Table~\ref{tab:data_stats}, we obtain 1,331 verified preference pairs across five capability levels as the final \textsc{LegalRewardBench}, with each sample consisting of a quadruplet with (\textit{query}, \textit{chosen response}, \textit{rejected response}, \textit{reference answer})\footnote{Representative data samples are shown in Appendix~\ref{app:benchmark_examples}.}. The substantial length of queries and responses demands that reward models perform robust comparisons under long-context and information-dense conditions.

To assess annotation reliability, we conduct an additional double-blind check on 5\% of the preference pairs, as shown in Table~\ref{tab:data_stats}. As can be seen, both double-blind human agreement and human agreement with automatic annotation reach an agreement around 80\%, verifying the effectiveness of the rubric-guided annotation scheme.

\begin{figure}[t]
\centering
\includegraphics[width=\linewidth]{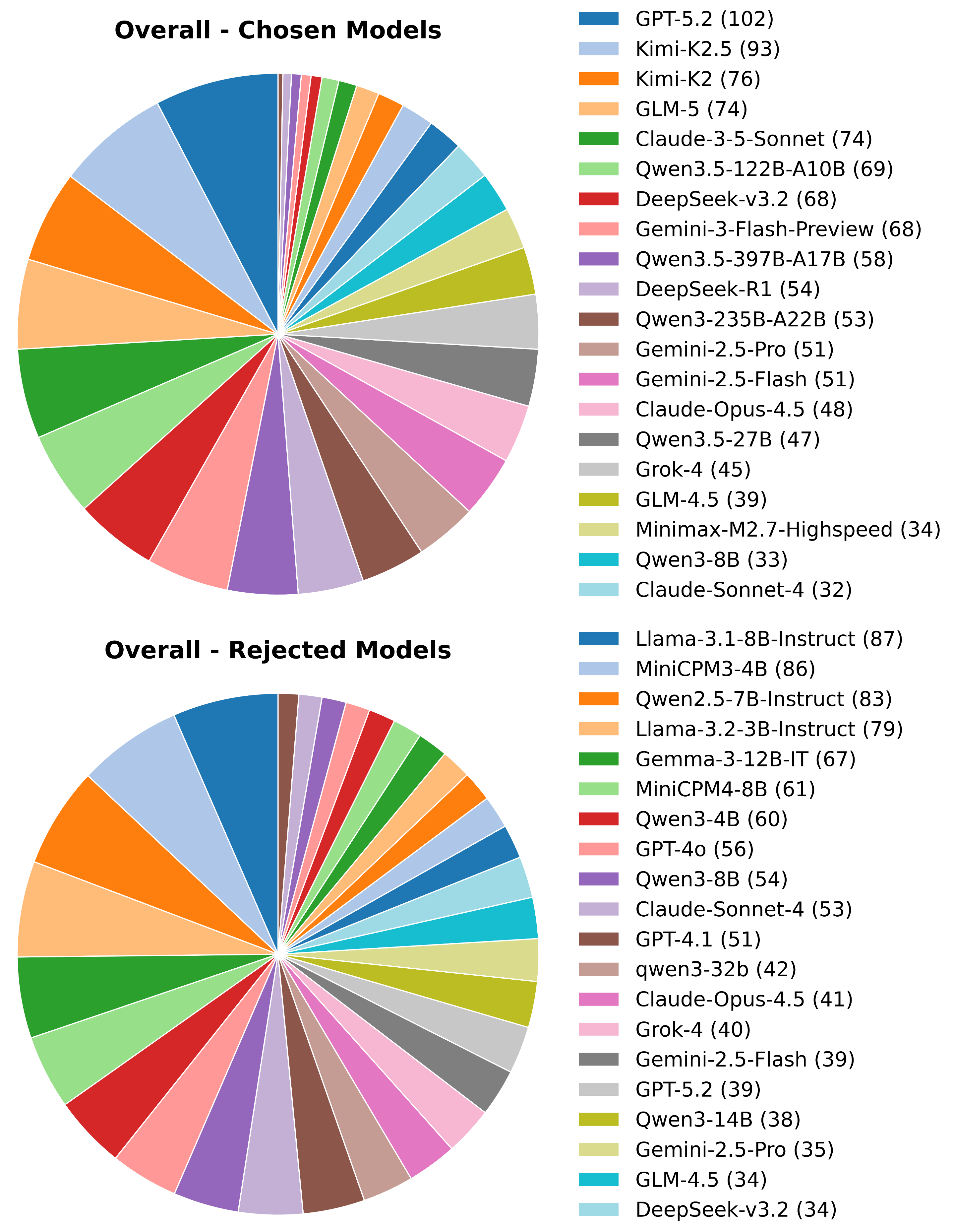}
\caption{Model distribution of candidate responses.}
\label{fig:response_dist}
\end{figure}

We further analyze the model distribution of chosen and rejected responses in Figure~\ref{fig:response_dist}. As can be seen, strong closed-source models account for a larger share of chosen responses, while medium-scale models appear more often among rejected responses. At the same time, the overall model distribution remains diverse, indicating the comprehensiveness of the responses.


\section{Benchmarking}


\subsection{Evaluation Setup}
\label{sec:experiment-setting}

We conduct evaluation on \textsc{LegalRewardBench} with three families of RMs\footnote{The detailed inference settings for all three RM families during our main experiment are presented in Appendix~\ref{app:inference_settings}.}:

\begin{enumerate}[leftmargin=4mm, itemsep=1mm, parsep=0pt]
    \item \textbf{General-purpose Generative Reward Models (General GenRM)}: These refer to general LLMs (e.g., Qwen-3 \cite{yang2025qwen3}) that perform evaluation via zero-shot prompting. We apply a standardized prompt template across this category to ensure a fair comparison.
    \item \textbf{Fine-tuned Generative Reward Models (Fine-tuned GenRM)}: These models are explicitly fine-tuned to produce textual judgment or scores such as \citet{mahan2024generative}. We evaluate them using their native templates to align with their specific fine-tuning designs.
    \item \textbf{Fine-tuned Discriminative Reward Models (Fine-tuned DisRM)}: These models are trained to output scalar scores with a newly initialized classification head such as \citet{liu2024skywork}. They typically process instruction-response pairs directly without evaluation prompts.
\end{enumerate}

We use accuracy on pairwise preference prediction as the primary metric\footnote{A self-bias analysis is presented in Appendix~\ref{app:self_bias}.}. To mitigate position bias \cite{shi2025judging,ZHOU2027104801} for generative models, we evaluate each pair twice with swapped response order and average the two results.

\subsection{Main Results}

\begin{table*}[t]
\centering
\small
\resizebox{\textwidth}{!}{
\begin{tabular}{llc|cccccc}
\toprule
\textbf{Model} & \textbf{Type} & \textbf{TK} & \textbf{Overall} & \textbf{L1} & \textbf{L2} & \textbf{L3} & \textbf{L4} & \textbf{L5} \\
\midrule
\rowcolor[HTML]{E0FFE0} Claude-Opus-4.7 & General LLM & \cmark & 86.85 & 91.83 & 81.36 & 87.00 & 87.50 & 82.66 \\
\rowcolor[HTML]{E0FFE0} Claude-Opus-4.5 & General LLM & \cmark & 85.01 & 91.01 & 76.75 & 86.60 & 84.12 & 82.66 \\
\rowcolor[HTML]{E0FFE0} GPT-5.5 & General LLM & \cmark & 83.66 & 92.97 & 78.73 & 88.06 & 72.97 & 81.85 \\
\rowcolor[HTML]{E0FFE0} Gemini-3-Flash-Preview & General LLM & \cmark & 82.95 & 89.22 & 74.12 & 84.88 & 82.09 & 79.84 \\
\rowcolor[HTML]{E0FFE0} \mQwenThreeTwoThreeFive & General LLM & \cmark & 81.52 & 88.40 & 73.25 & 84.08 & 77.87 & 80.65 \\
\rowcolor[HTML]{E0FFE0} Gemini-2.5-Pro & General LLM & \xmark & 81.14 & 88.40 & 75.22 & 83.55 & 79.56 & 70.56 \\
\rowcolor[HTML]{E0FFE0} \mKimiKTwoFive & General LLM & \cmark & 80.65 & 87.91 & 71.27 & 83.82 & 80.07 & 71.77 \\
\rowcolor[HTML]{E0FFE0} GPT-5.2 & General LLM & \cmark & 80.54 & 84.15 & 75.44 & 83.02 & 78.89 & 77.42 \\
\rowcolor[HTML]{E0FFE0} GLM-4.5 & General LLM & \cmark & 79.90 & 82.19 & 71.93 & 83.95 & 79.73 & 77.02 \\
\rowcolor[HTML]{E0FFE0} \mQwenThreeThirty & General LLM & \cmark & 79.30 & 78.92 & 74.78 & 82.49 & 79.90 & 77.42 \\
\rowcolor[HTML]{E0FFE0} GPT-4.1 & General LLM & \xmark & 78.96 & 77.94 & 72.15 & 83.42 & 83.11 & 70.56 \\
\rowcolor[HTML]{E0FFE0} \mQwenThreeFiveThreeNineSeven & General LLM & \cmark & 78.55 & 84.80 & 73.68 & 80.77 & 78.38 & 65.73 \\
\rowcolor[HTML]{E0FFE0} Gemini-2.5-Flash & General LLM & \cmark & 78.51 & 81.54 & 73.68 & 83.42 & 76.01 & 70.97 \\
\rowcolor[HTML]{E0FFE0} \mDeepSeekROne & General LLM & \cmark & 78.29 & 80.23 & 73.25 & 82.76 & 77.53 & 70.97 \\
\rowcolor[HTML]{FFFFCC} \mQwenNemotron & Ft. GenRM & \dashmark & 78.25 & 78.43 & 73.46 & 83.16 & 76.52 & 75.81 \\
\rowcolor[HTML]{E6F2FF} \mSkyworkQwenThree & Ft. DisRM & \dashmark & 78.14 & 77.12 & 70.61 & 84.08 & 79.73 & 72.58 \\
\rowcolor[HTML]{E0FFE0} \mDeepSeekVThreeTwo & General LLM & \cmark & 77.84 & 80.39 & 71.05 & 80.77 & 77.70 & 75.40 \\
\rowcolor[HTML]{E0FFE0} \mQwenThreeFourteen & General LLM & \cmark & 77.84 & 76.63 & 70.39 & 85.01 & 76.69 & 75.40 \\
\rowcolor[HTML]{E0FFE0} \mQwenThreeEight & General LLM & \cmark & 75.69 & 76.47 & 68.86 & 79.97 & 76.86 & 70.56 \\
\rowcolor[HTML]{E0FFE0} \mQwenThreeFiveTwentySeven & General LLM & \cmark & 75.69 & 81.70 & 65.35 & 81.17 & 75.84 & 62.90 \\
\rowcolor[HTML]{E0FFE0} Claude-Sonnet-4 & General LLM & \xmark & 75.51 & 77.61 & 71.71 & 78.91 & 72.80 & 73.39 \\
\rowcolor[HTML]{E0FFE0} \mQwenThreeFour & General LLM & \cmark & 74.87 & 71.90 & 70.18 & 82.23 & 74.66 & 68.95 \\
\rowcolor[HTML]{E0FFE0} \mLegalOne & General LLM & \xmark & 74.68 & 83.50 & 65.79 & 80.11 & 73.82 & 54.84 \\
\rowcolor[HTML]{E0FFE0} GPT-4o & General LLM & \xmark & 74.61 & 73.53 & 69.74 & 79.18 & 76.18 & 68.55 \\
\rowcolor[HTML]{E0FFE0} \mKimiKTwo & General LLM & \cmark & 74.08 & 77.61 & 65.35 & 78.65 & 72.97 & 70.16 \\
\rowcolor[HTML]{E0FFE0} \mGemmaThreeTwelve & General LLM & \xmark & 72.35 & 69.61 & 68.42 & 78.65 & 72.80 & 66.13 \\
\rowcolor[HTML]{E6F2FF} \mSkyworkLlamaThreeTwo & Ft. DisRM & \dashmark & 71.37 & 65.69 & 71.49 & 77.19 & 71.62 & 66.94 \\
\rowcolor[HTML]{E6F2FF} \mSkyworkLlamaThreeOne & Ft. DisRM & \dashmark & 70.85 & 68.30 & 64.91 & 75.60 & 75.00 & 63.71 \\
\rowcolor[HTML]{FFFFCC} \mRRMSeven & Ft. GenRM & \dashmark & 69.20 & 64.87 & 63.60 & 75.07 & 73.14 & 62.90 \\
\rowcolor[HTML]{E6F2FF} \mLlamaRM & Ft. DisRM & \dashmark & 68.90 & 60.46 & 67.11 & 75.07 & 72.97 & 64.52 \\
\rowcolor[HTML]{FFFFCC} \mRMRQwen & Ft. GenRM & \dashmark & 68.22 & 66.67 & 66.45 & 69.10 & 70.61 & 66.94 \\
\rowcolor[HTML]{E0FFE0} \mQwenThreeOneSeven & General LLM & \xmark & 67.66 & 63.24 & 61.40 & 71.22 & 74.49 & 62.90 \\
\rowcolor[HTML]{E0FFE0} \mLlamaThreeOne & General LLM & \xmark & 67.36 & 68.30 & 64.69 & 68.30 & 68.07 & 65.32 \\
\rowcolor[HTML]{E6F2FF} \mSkyworkLlamaVZeroTwo & Ft. DisRM & \dashmark & 67.02 & 61.44 & 62.28 & 67.90 & 72.64 & 73.39 \\
\rowcolor[HTML]{E0FFE0} \mQwenTwoFiveSeven & General LLM & \xmark & 65.78 & 62.58 & 60.75 & 67.11 & 69.59 & 69.76 \\
\rowcolor[HTML]{E6F2FF} \mInternLMReward & Ft. DisRM & \dashmark & 62.58 & 58.82 & 64.04 & 68.44 & 63.18 & 50.00 \\
\rowcolor[HTML]{E0FFE0} \mLlamaThreeTwo & General LLM & \xmark & 56.20 & 60.29 & 58.11 & 55.97 & 56.93 & 41.53 \\
\rowcolor[HTML]{E0FFE0} \mQwenThreeZeroSix & General LLM & \cmark & 55.41 & 56.21 & 51.97 & 50.13 & 63.51 & 56.45 \\
\bottomrule
\end{tabular}}
\caption{Main results on \textsc{LegalRewardBench}. Values are accuracies in percentage points.}
\label{tab:main_results}
\end{table*}

\begin{table*}[t]
\centering
\small
\resizebox{\textwidth}{!}{
\begin{tabular}{lcccccccccccc}
\toprule
\multirow{2}{*}{\textbf{Model}} & \multicolumn{2}{c}{\textbf{Overall}} & \multicolumn{2}{c}{\textbf{L1}} & \multicolumn{2}{c}{\textbf{L2}} & \multicolumn{2}{c}{\textbf{L3}} & \multicolumn{2}{c}{\textbf{L4}} & \multicolumn{2}{c}{\textbf{L5}} \\
\cmidrule(lr){2-3} \cmidrule(lr){4-5} \cmidrule(lr){6-7} \cmidrule(lr){8-9} \cmidrule(lr){10-11} \cmidrule(lr){12-13}
 & \textbf{Ori} & \textbf{Ref} & \textbf{Ori} & \textbf{Ref} & \textbf{Ori} & \textbf{Ref} & \textbf{Ori} & \textbf{Ref} & \textbf{Ori} & \textbf{Ref} & \textbf{Ori} & \textbf{Ref} \\
\midrule
Claude-Opus-4.7 & 86.85 & \textbf{91.81} & 91.83 & \textbf{94.93} & 81.36 & \textbf{90.13} & 87.00 & \textbf{88.99} & 87.50 & \textbf{96.11} & 82.66 & \textbf{85.48} \\
GPT-4o & 74.61 & \textbf{79.30} & 73.53 & \textbf{84.15} & 69.74 & \textbf{76.10} & 79.18 & \textbf{79.97} & 76.18 & \textbf{81.76} & \textbf{68.55} & 65.32 \\
GPT-5.2 & 80.54 & \textbf{88.73} & 84.15 & \textbf{90.52} & 75.44 & \textbf{84.43} & 83.02 & \textbf{87.53} & 78.89 & \textbf{93.92} & 77.42 & \textbf{83.47} \\
\mKimiKTwoFive & 80.65 & \textbf{83.43} & 87.91 & \textbf{89.05} & 71.27 & \textbf{73.90} & 83.82 & \textbf{85.68} & 80.07 & \textbf{89.19} & \textbf{71.77} & 66.53 \\
\mQwenThreeFourteen & 77.84 & \textbf{86.14} & 76.63 & \textbf{90.69} & 70.39 & \textbf{82.46} & 85.01 & \textbf{87.80} & 76.69 & \textbf{85.47} & 75.40 & \textbf{78.23} \\
\bottomrule
\end{tabular}}
\caption{Averaged accuracy on \textsc{LegalRewardBench} with reference included or excluded.}
\label{tab:reference}
\end{table*}

\begin{table*}[!t]
\centering
\small
\resizebox{\textwidth}{!}{
\begin{tabular}{lcccccccccccc}
\toprule
\multirow{2}{*}{\textbf{Model}} & \multicolumn{2}{c}{\textbf{Overall}} & \multicolumn{2}{c}{\textbf{L1}} & \multicolumn{2}{c}{\textbf{L2}} & \multicolumn{2}{c}{\textbf{L3}} & \multicolumn{2}{c}{\textbf{L4}} & \multicolumn{2}{c}{\textbf{L5}} \\
\cmidrule(lr){2-3} \cmidrule(lr){4-5} \cmidrule(lr){6-7} \cmidrule(lr){8-9} \cmidrule(lr){10-11} \cmidrule(lr){12-13}
 & \textbf{Tk} & \textbf{NoTk} & \textbf{Tk} & \textbf{NoTk} & \textbf{Tk} & \textbf{NoTk} & \textbf{Tk} & \textbf{NoTk} & \textbf{Tk} & \textbf{NoTk} & \textbf{Tk} & \textbf{NoTk} \\
\midrule
Claude-Opus-4.7 & \textbf{86.85} & 86.18 & \textbf{91.83} & 90.52 & 81.36 & \textbf{81.58} & \textbf{87.00} & 86.60 & \textbf{87.50} & 86.66 & \textbf{82.66} & 81.45 \\
Gemini-3-Flash & \textbf{82.95} & 82.34 & \textbf{89.22} & 88.73 & \textbf{74.12} & 73.25 & \textbf{84.88} & 84.22 & 82.09 & \textbf{82.26} & \textbf{79.84} & 77.82 \\
\mQwenThreeFourteen & \textbf{77.84} & 76.97 & \textbf{76.63} & 74.67 & 70.39 & \textbf{72.15} & \textbf{85.01} & 80.64 & 76.69 & \textbf{77.53} & 75.40 & \textbf{79.03} \\
\mQwenThreeFour & \textbf{74.87} & 73.14 & \textbf{71.90} & 70.26 & \textbf{70.18} & 69.08 & \textbf{82.23} & 81.70 & \textbf{74.66} & 72.30 & \textbf{68.95} & 63.71 \\
\mQwenThreeEight & 75.69 & \textbf{76.60} & \textbf{76.47} & 75.49 & 68.86 & \textbf{71.93} & 79.97 & \textbf{82.49} & \textbf{76.86} & 76.01 & 70.56 & \textbf{71.37} \\
\bottomrule
\end{tabular}}
\caption{Averaged accuracy on \textsc{LegalRewardBench} with thinking mode on or off.}
\label{tab:think}
\end{table*}

\begin{table*}[!t]
\centering
\small
\resizebox{\textwidth}{!}{
\begin{tabular}{lcccccccccccc}
\toprule
\multirow{2}{*}{\textbf{Model}} & \multicolumn{2}{c}{\textbf{Overall}} & \multicolumn{2}{c}{\textbf{L1}} & \multicolumn{2}{c}{\textbf{L2}} & \multicolumn{2}{c}{\textbf{L3}} & \multicolumn{2}{c}{\textbf{L4}} & \multicolumn{2}{c}{\textbf{L5}} \\
\cmidrule(lr){2-3} \cmidrule(lr){4-5} \cmidrule(lr){6-7} \cmidrule(lr){8-9} \cmidrule(lr){10-11} \cmidrule(lr){12-13}
 & \textbf{Zh} & \textbf{En} & \textbf{Zh} & \textbf{En} & \textbf{Zh} & \textbf{En} & \textbf{Zh} & \textbf{En} & \textbf{Zh} & \textbf{En} & \textbf{Zh} & \textbf{En} \\
\midrule
Claude-Opus-4.7 & \textbf{86.85} & 85.54 & \textbf{91.83} & 89.71 & 81.36 & \textbf{81.58} & \textbf{87.00} & 85.94 & \textbf{87.50} & 84.97 & \textbf{82.66} & \textbf{82.66} \\
GPT-4o & 74.61 & \textbf{74.98} & \textbf{73.53} & 71.73 & 69.74 & \textbf{70.61} & 79.18 & \textbf{79.44} & 76.18 & \textbf{79.05} & \textbf{68.55} & 67.74 \\
GPT-5.2 & \textbf{80.54} & 80.09 & \textbf{84.15} & 83.82 & 75.44 & \textbf{75.66} & \textbf{83.02} & 82.49 & \textbf{78.89} & 77.36 & 77.42 & \textbf{78.23} \\
\mKimiKTwoFive & \textbf{80.65} & 78.36 & \textbf{87.91} & 85.46 & \textbf{71.27} & 69.52 & \textbf{83.82} & 80.77 & \textbf{80.07} & 79.22 & \textbf{71.77} & 67.74 \\
\mQwenThreeFourteen & 77.84 & \textbf{78.44} & 76.63 & \textbf{77.78} & 70.39 & \textbf{74.12} & 85.01 & \textbf{85.28} & \textbf{76.69} & 76.18 & \textbf{75.40} & 72.58 \\
\bottomrule
\end{tabular}}
\caption{Averaged accuracy on \textsc{LegalRewardBench} with different thinking languages.}
\label{tab:cot_language}
\end{table*}

\begin{table*}[!t]
\centering
\small
\resizebox{\textwidth}{!}{
\begin{tabular}{lcccccccccccc}
\toprule
\multirow{2}{*}{\textbf{Model}} & \multicolumn{2}{c}{\textbf{Overall}} & \multicolumn{2}{c}{\textbf{L1}} & \multicolumn{2}{c}{\textbf{L2}} & \multicolumn{2}{c}{\textbf{L3}} & \multicolumn{2}{c}{\textbf{L4}} & \multicolumn{2}{c}{\textbf{L5}} \\
\cmidrule(lr){2-3} \cmidrule(lr){4-5} \cmidrule(lr){6-7} \cmidrule(lr){8-9} \cmidrule(lr){10-11} \cmidrule(lr){12-13}
 & \textbf{Ori} & \textbf{Rub} & \textbf{Ori} & \textbf{Rub} & \textbf{Ori} & \textbf{Rub} & \textbf{Ori} & \textbf{Rub} & \textbf{Ori} & \textbf{Rub} & \textbf{Ori} & \textbf{Rub} \\
\midrule
Claude-Opus-4.7 & \textbf{86.85} & 85.99 & \textbf{91.83} & 87.75 & \textbf{81.36} & 80.48 & 87.00 & \textbf{87.14} & 87.50 & \textbf{88.85} & \textbf{82.66} & 81.45 \\
GPT-4o & 74.61 & \textbf{75.88} & \textbf{73.53} & 72.55 & 69.74 & \textbf{71.49} & 79.18 & \textbf{81.70} & 76.18 & \textbf{79.22} & \textbf{68.55} & 66.53 \\
GPT-5.2 & 80.54 & \textbf{80.73} & \textbf{84.15} & 82.19 & 75.44 & \textbf{76.32} & \textbf{83.02} & 81.56 & 78.89 & \textbf{81.76} & 77.42 & \textbf{80.24} \\
\mKimiKTwoFive & \textbf{80.65} & 71.83 & \textbf{87.91} & 79.74 & \textbf{71.27} & 62.94 & \textbf{83.82} & 74.40 & \textbf{80.07} & 75.00 & \textbf{71.77} & 53.23 \\
\mQwenThreeFourteen & 77.84 & \textbf{78.21} & \textbf{76.63} & 76.14 & 70.39 & \textbf{73.46} & \textbf{85.01} & 82.89 & 76.69 & \textbf{78.55} & 75.40 & \textbf{77.02} \\
\bottomrule
\end{tabular}}
\caption{Averaged accuracy on \textsc{LegalRewardBench} with automatic rubric generation.}
\label{tab:rubric}
\end{table*}

Table~\ref{tab:main_results} reports the results on \textsc{LegalRewardBench}. We summarize our findings as follows:

\paragraph{Finding 1: General GenRMs substantially outperform most fine-tuned reward models.} Top-performing models are mainly from the Claude, GPT, Gemini, and Qwen3 families, where the most updated Claude-Opus-4.7 achieves nearly 87 points. On the other hand, fine-tuned generative and discriminative RMs perform more modestly, and neither family clearly dominates the other. This differs from previous RM benchmarks, where strong discriminative RMs can achieve top performance \cite{lambert2025rewardbench,malik2025rewardbench}. This likely stems from the lack of high-quality legal preference data, while general-domain preferences hardly transfer to the legal domain.

\paragraph{Finding 2: Legal information extraction and generation remain difficult for legal preference annotation.} RMs perform comparatively better on \texttt{knowledge} and \texttt{analysis} tasks, likely benefiting from the knowledge and reasoning abilities of recent LLMs \cite{sun2025survey}. On the other hand, RMs struggle on \texttt{understanding} and \texttt{writing} related tasks. This demonstrates that the current models still struggle on information extraction and generation tasks, which cannot be easily improved by scaled model parameters or reasoning length.

\paragraph{Finding 3: Legal-specialized LLMs do not necessarily become strong legal RMs.}LegalOne-8B \cite{li2026legalone}, for example, does not show a clear advantage in this benchmark. This suggests that solving legal tasks and evaluating legal responses are not necessarily related capabilities.

\subsection{Analysis}

\subsubsection{Effect of Reference Answers}

As detailed in Section 2, we curate a reference answer for every query in our benchmark. To evaluate the impact of these references, we compare the baseline setting (where the RM receives only the query and candidate responses) against the reference-augmented setting (where the reference is included in the input), as shown in Table~\ref{tab:reference}.

\paragraph{Finding 4: Reference answers significantly improve legal reward modeling.} This improvement is particularly pronounced in \texttt{knowledge}, \texttt{understanding}, and \texttt{decision} tasks. This aligns with the inherent characteristics of the legal tasks: unlike general open-ended tasks that lack a single standard response, legal queries are typically grounded in specific statutes and precedents, leading to more determinate reference answers. Therefore, reference answers should be leveraged whenever available in legal reward modeling.

\subsubsection{Effect of Thinking Enhancement}

Legal tasks inherently require extensive reasoning. We therefore compare \textit{Think} and \textit{No Think} settings for representative models while keeping other settings fixed\footnote{For hybrid reasoning models, we toggle the reasoning mode on or off; for other models, we compare the reasoning-enabled versions with their standard counterparts.}. We also test whether the language of reasoning process affects results by explicitly specifying the reasoning language. The results are presented in Tables~\ref{tab:think} and~\ref{tab:cot_language}.

\paragraph{Finding 5: Thinking enhancement usually improves performance on \textsc{LegalRewardBench}.} This suggests that explicit deliberation helps the RM compare legal reasoning, detect unsupported claims, and avoid being misled by surface fluency. On the other hand, no consistent advantage is seen for thinking in a specific language. Although the dataset is in Chinese, legal reward modeling appears to depend more on the precision of the thinking process rather than the thinking language.

\subsubsection{Effect of Rubric Generation}

As discussed in Section \ref{sec:annotation}, the integration of rubrics is instrumental in enhancing annotation consistency.
Therefore, we further investigate whether autonomous rubric generation can also improve RM performance. Specifically, we ask the generative LLMs to first write a brief rubric for each query and then compare the candidate responses based on the rubric. This turns the evaluation into a two-step process. The results are in Table~\ref{tab:rubric}.

\paragraph{Finding 6: Model-generated rubrics do not consistently improve performance.} A likely explanation is that human-written legal rubrics are highly task-specific, while model-generated rubrics tend to remain generic, emphasizing broad criteria such as clarity and completeness. Such generic rubrics may fail to capture concrete legal issues. Despite the potential of rubrics to further improve RM accuracy \cite{gunjal2025rubrics}, it is non-trivial to obtain effective rubrics for legal reward modeling. 

\subsubsection{\texorpdfstring{Best-of-$n$ Validation}{Best-of-n Validation}}

\begin{table*}[!t]
\centering
\small
\resizebox{\textwidth}{!}{
\begin{tabular}{ccllccc}
\toprule
\textbf{Rank} & \textbf{Ori\_Rank} & \textbf{Base Model} & \textbf{Sampler} & \textbf{Average} & \textbf{LawBench} & \textbf{LexEval} \\
\midrule
1 & 1 & Qwen3-4B & Claude-Opus-4.5 & 54.29 & 52.12 & 56.46 \\
2 & 3 & Qwen3-4B & GPT-5.2 & 54.14 & 51.26 & 57.02 \\
3 & 2 & Qwen3-4B & Gemini-3-Flash-Preview & 54.00 & 51.26 & 56.74 \\
4 & 4 & Qwen3-4B & GPT-4.1 & 52.70 & 49.65 & 55.75 \\
5 & 5 & Qwen3-4B & Skywork-Reward-V2-Qwen3-8B & 52.38 & 49.23 & 55.52 \\
6 & 7 & Qwen3-4B & GPT-4o & 52.19 & 49.54 & 54.84 \\
7 & 6 & Qwen3-4B & Qwen3-8B & 51.87 & 49.43 & 54.32 \\
8 & 8 & Qwen3-4B & Skywork-Reward-V2-Llama-3.2-3B & 51.47 & 48.27 & 54.67 \\
9 & 9 & Qwen3-4B & Skywork-Reward-V2-Llama-3.1-8B & 51.28 & 48.68 & 53.88 \\
-- & -- & Qwen3-4B & Best-of-1 Baseline & 51.02 & 48.19 & 53.85 \\
10 & 10 & Qwen3-4B & Skywork-Reward-Llama-3.1-8B-v0.2 & 50.94 & 49.21 & 52.67 \\
\bottomrule
\end{tabular}}
\caption{Best-of-4 sampling validation on LawBench and LexEval, with Qwen3-4B as the base model and different RMs as the sampler. \textit{Ori\_Rank} denotes the RM's original rank on \textsc{LegalRewardBench}. }
\label{tab:bon}
\end{table*}

We further evaluate whether scores on \textsc{LegalRewardBench} serve as a reliable predictor for downstream performance gains when an RM is employed for model alignment. Specifically, we conduct a Best-of-$n$ validation on LawBench \cite{fei2024lawbench} and LexEval \cite{li2024lexeval}. We construct a 50-sample subset for each external benchmark and compare representative RMs in a best-of-4 setting due to computational constraints. The overall results are reported in Table~\ref{tab:bon}.

\paragraph{Finding 7: Performance on \textsc{LegalRewardBench} strongly correlates with downstream sampling effectiveness.} The trends are highly consistent across both benchmarks, with a Spearman correlation of 0.9758 between \textsc{LegalRewardBench} ranking and downstream selection gains with the 10 selected RMs. This supports the practical validity of the benchmark: it measures differences among legal RMs that are strongly reflected in actual response selection.

\section{Model Training}
\label{sec:model_training}

Despite the underperformance of general-domain fine-tuned RMs on \textsc{LegalRewardBench}, in this section, we investigate whether we can train a legal-domain reward model that transfers more effectively to legal preference judgments.

\subsection{Training Setup}

To enable reliable, reference-based preference annotation, we prioritize legal queries accompanied by authoritative reference answers. Concretely, we aggregate raw data from two primary sources: legal evaluation data from LegalDelta~\cite{dai2025legal} and in-house judicial examination data.

We further implement a rigorous preprocessing pipeline: 1) Remove instances with low-quality queries and reference answers. 2) Filter out examples highly similar to \textsc{LegalRewardBench} to prevent evaluation data leakage. 3) Rewrite all queries into a unified format that instructs the model to explicitly reason before answering. 4) Reduce the proportion of simplistic objective tasks such as multi-choice questions to less than 30\%.

Following preprocessing, we construct preference pairs from the cleaned query-reference dataset. For each query, we sample 6 responses from two distinct base models and evaluate them using GPT-5.2 under a reference-based scoring protocol\footnote{The detailed prompt template is presented in Prompt~\ref{prompt:training_data_scoring}.}. The highest- and lowest-scoring responses are designated as the chosen and rejected instances, respectively. This pipeline yields a final training set of approximately 39K high-quality preference pairs, consisting of 33K exam data-based pairs and 6K pairs derived from LegalDelta.

Using Qwen3-8B as the backbone, we compare two prominent RM training paradigms:
\begin{enumerate}[leftmargin=4mm, itemsep=1mm, parsep=0pt]
    \item \textbf{Discriminative RM:} This approach trains a sequence classifier with a scalar scoring head. Specifically, we incorporate Bradley-Terry objective \cite{liu2024skywork} to learn preference scoring based on pairwise data.
    \item \textbf{Generative RM:} This approach reformulates reward modeling as a generative preference optimization task \cite{huang2026think}. Specifically, we leverage a GRPO-based objective \cite{shao2024deepseekmath}, utilizing the correctness of the preference prediction as the reward signal.
\end{enumerate}

\begin{table*}[t]
\centering
\resizebox{\linewidth}{!}{
\begin{tabular}{cc cccccc cccccc}
\toprule
\multirow{2}{*}{\textbf{Method}} & \multirow{2}{*}{\textbf{Data}} & \multicolumn{6}{c}{\textbf{Qwen3-8B}} & \multicolumn{6}{c}{\textbf{Qwen2.5-7B-Instruct}} \\
\cmidrule(lr){3-8} \cmidrule(lr){9-14}
& & \textbf{L1} & \textbf{L2} & \textbf{L3} & \textbf{L4} & \textbf{L5} & \textbf{Overall} & \textbf{L1} & \textbf{L2} & \textbf{L3} & \textbf{L4} & \textbf{L5} & \textbf{Overall} \\
\midrule
Prompt & -- & 76.47 & 68.86 & 79.97 & 76.86 & 70.56 & 75.69 & 62.58 & 60.75 & 67.11 & 69.59 & 69.76 & 65.78 \\
w/ reference & -- & 88.42 & 78.98 & 83.38 & 82.61 & 69.01 & \textbf{82.26} & 76.31 & 64.91 & 70.03 & 72.64 & 69.35 & 71.11 \\
\midrule
\multirow{2}{*}{DisRM} & LegalDelta 6K & 72.55 & 60.96 & 65.78 & 73.65 & 67.74 & 68.44 & 76.14 & 73.25 & 73.47 & 73.31 & 62.90 & \textbf{73.03} \\
& Combined 39K & 78.43 & 64.91 & 72.41 & 77.03 & 66.13 & 72.95 & 73.86 & 64.47 & 74.01 & 76.35 & 70.97 & 72.58 \\
\midrule
\multirow{2}{*}{GenRM} & LegalDelta 6K & 74.02 & 69.74 & 82.49 & 77.87 & 69.76 & 76.15 & 65.69 & 56.58 & 62.07 & 72.13 & 58.06 & 63.82 \\
& Combined 39K & 74.51 & 76.10 & 83.69 & 81.25 & 75.40 & 78.96 & 70.10 & 68.20 & 74.27 & 71.45 & 65.32 & 70.81 \\
\bottomrule
\end{tabular}}
\caption{Comparison of different RM training methods on \textsc{LegalRewardBench}.}
\label{tab:model_training}
\end{table*}

\subsection{Results and Analysis}

\paragraph{Finding 8: It is non-trivial to fine-tune a legal-domain reward model.} As shown in Table~\ref{tab:model_training}, RM training brings limited and inconsistent gains. GenRM training yields only marginal improvements over the prompt-based judge, while DisRM training can even underperform it. These results suggest that simply converting reference-scored legal data into preference pairs is insufficient for cultivating a superior legal RM. Notably, the LegalDelta data was successfully applied for LLM post-training with substantial improvements~\cite{dai2025legal}, which largely eliminates poor query quality as the primary bottleneck for our RM's underperformance.

In summary, effective legal reward modeling extends far beyond merely aggregating domain-specific data. In contrast, simply introducing reference answers yields a performance boost of nearly 7 percentage points without the risk of compromising the model's inherent generalization capabilities \cite{huang2025empirical}. Therefore, for practical deployments, directly leveraging strong off-the-shelf generative models coupled with prompting techniques presents a more robust alternative.


\section{Related Work}

\subsection{Reward Model Benchmarks}

With the prevalence of RM-based reinforcement learning during LLM alignment process \cite{dong2024rlhfworkflowrewardmodeling}, systematic evaluation of reward models has grown rapidly in recent years. RewardBench~\citep{lambert2025rewardbench} established the first open evaluation framework for comparisons among different reward-model types. RewardBench 2~\citep{malik2025rewardbench} further introduced more difficult examples and broader skill coverage.

Beyond general RM evaluation, recent work has extended RM benchmarks to specialized settings. IF-RewardBench focuses on instruction-following evaluation \cite{wen2026if}; Multimodal RewardBench extends reward modeling to multimodal scenarios \cite{yasunaga2025multimodalrewardbenchholisticevaluation}; RAG-RewardBench studies multi-hop reasoning, citation faithfulness, and robustness in retrieval-augmented generation \cite{jin2024rag}; and Agent-RewardBench evaluates reward models in agentic tasks \cite{lù2025agentrewardbenchevaluatingautomaticevaluations}. These benchmarks show that RM evaluation is highly dependent on task and domain structure. In the legal domain, LRB~\citep{franzone2026buildinglegalrewardmodels} constructs retrieval-grounded preference pairs labeled solely by LLM judges, but a human-verified RM benchmark is still lacking.



\subsection{Reward Modeling in the Legal Domain}

As a critical application domain for natural language processing (NLP), various legal NLP benchmarks have been established. LawBench \cite{fei2024lawbench} serves as a comprehensive benchmark for assessing LLMs in the Chinese legal field, evaluating their ability from three dimensions. Similarly, LexEval \cite{li2024lexeval} provides a framework for measuring the cognitive and practical capabilities of LLMs across six dimensions. However, existing works primarily focus on whether a model can perform legal tasks instead of reward modeling.

Recent work has also explored reinforcement learning (RL) based alignment for legal models. LegalDelta improves legal reasoning through chain-of-thought-guided information gain and reward signals from verifiable legal tasks \cite{dai2025legal}. Unilaw-R1 conducts reinforcement learning combining format rewards, accuracy rewards, and legal-effect rewards \cite{cai2025unilaw}. These works demonstrate that RL-based legal post-training is useful and necessary, but the design of more effective RMs for legal alignment remains a gap.


\section{Conclusion}

In this work, we introduce \textsc{LegalRewardBench}, the first human-verified benchmark for reward modeling in the legal domain. Our evaluation shows that strong general-purpose LLMs are currently the most robust RMs for the legal domain, while existing fine-tuned RMs transfer less reliably to legal tasks, and it is non-trivial to fine-tune legal-domain RMs. Considering the lack of high-quality preference data and the knowledge- and reasoning-intensive nature of legal tasks, we strongly recommend using existing general models with knowledge and reasoning enhancement for legal reward modeling.


\section*{Limitations}

This work still has several limitations:

First, the current version mainly covers Chinese legal tasks due to practical constraints: high-quality legal annotation requires annotators who understand both the language and the relevant legal system, and it is difficult to obtain reliable annotation resources for multiple jurisdictions at the same time. Future versions should extend \textsc{LegalRewardBench} to more languages and legal systems.

Second, although we discuss preliminary directions for legal reward-model training, this paper does not introduce a new training method or report complete training results. Future work should investigate stronger legal reward-model systems, including knowledge-enhanced judges, higher-quality legal rubrics, reference-aware reward models, and training data designed specifically for legal preference judgment.

\section*{Ethics Considerations}

This work involves several ethical considerations:

\paragraph{Data and Consent}

Evaluation queries of the benchmark primarily originate from three sources: in-house legal task data, existing legal benchmarks, and historical judicial examination materials. The training phase additionally incorporates LegalDelta and in-house exam materials. We process public-source data strictly when permitted by the original platform or publisher for research, crawling, or redistribution, ensuring full compliance with respective data-use agreements, platform terms, and open-source licenses. For internal data, our research team ensures legitimate authorization and strictly controlled data circulation. 

Furthermore, we recognize copyright and licensing constraints, acknowledging that not all legal texts, examination materials, or benchmark data are exempt from reuse restrictions.

\paragraph{Human Annotators and Crowdsourcing}

The benchmark construction involves a rigorous human-verification stage where annotators with domain expertise participate in rubric creation and pairwise preference validation. All annotators were fully informed that their contributions would be used for academic research, ensuring clear comprehension of the task, workload, and data usage. Compensation was fairly commensurate with the time and effort required, strictly adhering to reasonable hourly wage standards to prevent exploitation. 

In addition, we acknowledge that human annotation inherently risks introducing subjectivity and institutional bias. Although we mitigate these risks via rubric-guided evaluation, hybrid automatic-human review, and double-blind agreement checks, we emphasize that human annotations represent constrained research preference signals rather than absolute, definitive legal judgments.

\paragraph{Potential Harms and Broader Impact}

The primary objective of this work is to benchmark reward models within the legal domain. However, advanced legal reward models could potentially be integrated into downstream applications for post-training or answer filtering. This poses a risk of misuse in high-stakes decision-making scenarios, potentially generating plausible yet legally unsound outputs that amplify user over-reliance. 

Additionally, bias and fairness remain unavoidable challenges. Legal corpora, historical exams, reference answers, rubrics, and annotator judgments may implicitly harbor institutional biases or stylistic inclinations. Therefore, we explicitly position this work as a rigorous academic evaluation tool, rather than a foundation for autonomous decision-making in real-world legal practices.

\bibliography{custom}
\clearpage
\appendix

\section{Model Catalog}
\label{app:model_catalog}

Table~\ref{tab:model_catalog} lists all models involved in \textsc{LegalRewardBench}, including both candidate response generation and reward modeling evaluation. For each model, we report its reward-model type when applicable, whether it is evaluated with thinking enhancement (TK), and whether it is used for response generation or reward-model evaluation. This catalog shows that our benchmark covers models from diverse families, architectures, and scales across both construction and benchmarking.

\begin{table*}[!h]
\centering
\scriptsize
\resizebox{0.85\linewidth}{!}{
\setlength{\tabcolsep}{4pt}
\renewcommand{\arraystretch}{0.97}
\begin{tabular}{llcccc}
\toprule
\textbf{Model} & \textbf{Type} & \textbf{TK} & \textbf{Size} & \textbf{Response Gen.} & \textbf{RM} \\
\midrule
\multicolumn{6}{l}{\textit{Open-weight General LLMs}} \\
\mQwenThreeFiveThreeNineSeven & General LLM & \cmark & 397B (17B) & \cmark & \cmark \\
Qwen3.5-122B-A10B & General LLM & \cmark & 122B (10B) & \cmark & \xmark \\
\mQwenThreeFiveTwentySeven & General LLM & \cmark & 27B & \cmark & \cmark \\
\mQwenThreeTwoThreeFive & General LLM & \cmark & 235B (22B) & \cmark & \cmark \\
Qwen3-32B & General LLM & \cmark & 32B & \cmark & \xmark \\
\mQwenThreeThirty & General LLM & \cmark & 30B (3B) & \xmark & \cmark \\
\mQwenThreeFourteen & General LLM & \cmark & 14B & \cmark & \cmark \\
\mQwenThreeEight & General LLM & \cmark & 8B & \cmark & \cmark \\
\mQwenThreeFour & General LLM & \cmark & 4B & \cmark & \cmark \\
\mQwenThreeOneSeven & General LLM & \cmark & 1.7B & \xmark & \cmark \\
\mQwenThreeZeroSix & General LLM & \cmark & 0.6B & \xmark & \cmark \\
\mQwenTwoFiveSeven & General LLM & \xmark & 7B & \cmark & \cmark \\
\mDeepSeekVThreeTwo & General LLM & \cmark & 671B (37B) & \cmark & \cmark \\
\mDeepSeekROne & General LLM & \cmark & 671B (37B) & \cmark & \cmark \\
\mKimiKTwoFive & General LLM & \cmark & 1T (32B) & \cmark & \cmark \\
\mKimiKTwo & General LLM & \cmark & 1T (32B) & \cmark & \cmark \\
GLM-5 & General LLM & \cmark & 744B (40B) & \cmark & \xmark \\
GLM-4.5 & General LLM & \cmark & 355B (32B) & \cmark & \cmark \\
\mMiniCPMFour & General LLM & \xmark & 8B & \cmark & \cmark \\
MiniCPM3-4B & General LLM & \xmark & 4B & \cmark & \xmark \\
Minimax-M2.7 & General LLM & \cmark & 230B (10B) & \cmark & \xmark \\
\mGemmaThreeTwelve & General LLM & \xmark & 12B & \cmark & \cmark \\
\mLegalOne & General LLM & \xmark & 8B & \xmark & \cmark \\
\mLlamaThreeOne & General LLM & \xmark & 8B & \cmark & \cmark \\
\mLlamaThreeTwo & General LLM & \xmark & 3B & \cmark & \cmark \\
\midrule
\multicolumn{6}{l}{\textit{Closed-source General LLMs}} \\
Claude-Opus-4.7 & General LLM & \cmark & \dashmark & \xmark & \cmark \\
Claude-Opus-4.5 & General LLM & \cmark & \dashmark & \cmark & \cmark \\
Claude-Sonnet-4 & General LLM & \xmark & \dashmark & \cmark & \cmark \\
Claude-3-5-Sonnet & General LLM & \xmark & \dashmark & \cmark & \xmark \\
Gemini-3-Flash-Preview & General LLM & \cmark & \dashmark & \cmark & \cmark \\
Gemini-2.5-Pro & General LLM & \xmark & \dashmark & \cmark & \cmark \\
Gemini-2.5-Flash & General LLM & \cmark & \dashmark & \cmark & \cmark \\
GPT-5.5 & General LLM & \cmark & \dashmark & \cmark & \cmark \\
GPT-5.2 & General LLM & \cmark & \dashmark & \cmark & \cmark \\
GPT-4.1 & General LLM & \xmark & \dashmark & \cmark & \cmark \\
GPT-4o & General LLM & \xmark & \dashmark & \cmark & \cmark \\
Grok-4 & General LLM & \xmark & \dashmark & \cmark & \xmark \\
\midrule
\multicolumn{6}{l}{\textit{Fine-tuned generative reward models}} \\
\mQwenNemotron & Ft. GenRM & \dashmark & 8B & \xmark & \cmark \\
\mRMRQwen & Ft. GenRM & \dashmark & 7B & \xmark & \cmark \\
\mRRMSeven & Ft. GenRM & \dashmark & 7B & \xmark & \cmark \\
\midrule
\multicolumn{6}{l}{\textit{Fine-tuned discriminative reward models}} \\
\mSkyworkQwenThree & Ft. DisRM & \dashmark & 8B & \xmark & \cmark \\
\mSkyworkLlamaThreeOne & Ft. DisRM & \dashmark & 8B & \xmark & \cmark \\
\mLlamaRM & Ft. DisRM & \dashmark & 8B & \xmark & \cmark \\
\mSkyworkLlamaVZeroTwo & Ft. DisRM & \dashmark & 8B & \xmark & \cmark \\
\mSkyworkLlamaThreeTwo & Ft. DisRM & \dashmark & 3B & \xmark & \cmark \\
\mInternLMReward & Ft. DisRM & \dashmark & 1.8B & \xmark & \cmark \\
\bottomrule
\end{tabular}}
\caption{Model catalog for \textsc{LegalRewardBench}. TK denotes whether the model is evaluated in thinking mode; \dashmark{} indicates that the distinction is not applicable to the model family.}
\label{tab:model_catalog}
\end{table*}

\section{Task Distribution}
\label{app:task_description}
Table~\ref{tab:task_description} summarizes the task composition of \textsc{LegalRewardBench}. For each level, we list the task name, sample count in the final benchmark, data source, and scoring type. The final benchmark contains 41 task types in 5 capability levels.

\section{Annotation Details}
\label{app:annotation_guidelines}

As explained in Section~\ref{sec:annotation}, our rubric-based annotation pipeline consists of two stages: automatic scoring and human verification.

For automatic scoring, we use two scoring protocols. For tasks that can be easily verified, we use rule-based scoring. Specifically, for statute recitation task, we compute ROUGE-L for the final output and the reference answer based on character-level segmentation, and map the score to a 0--100 scale. For the two sentence-prediction tasks, we start from 100 points and deduct one point for each month of absolute difference between the predicted and reference sentence terms, with a lower bound of 0. All remaining tasks use model-based scoring with Gemini-3.1-Pro-Preview as the LLM judge, and the automatic-scoring prompt (Prompt~\ref{prompt:auto_scoring}) asks an LLM judge to assign an integer score to each candidate response according to task-specific rubrics. This design separates tasks with deterministic reference-based metrics from open-ended legal reasoning or writing tasks that require rubric-based model judging.

For human verification, we hired 5 Chinese master's students in law. The human annotation prompts in Prompt~\ref{prompt:annotation_short} and Prompt~\ref{prompt:annotation_long} then ask annotators to verify pairwise preferences, using the rubrics as the primary comparison criterion while still checking for clear legal-quality issues. For L5 \texttt{Writing} tasks, we additionally provide the reasoning produced by the automatic scorer as auxiliary evidence, because these queries often contain extensive background materials and are more difficult for annotators to compare directly. The whole process of human verification takes 560 human-hours.

\newcommand{\taskcell}[2]{\parbox[c]{2.10in}{\raggedright #1\\{\footnotesize (#2)}}}
\newcommand{\sourcecell}[2]{\parbox[c]{1.10in}{\centering #1\\(#2)}}
\newcommand{\levelcell}[2]{\parbox[c]{0.90in}{\centering #1\\{\footnotesize(\texttt{#2})}}}

\begin{table*}[!h]
\centering
\small
\begin{CJK*}{UTF8}{gbsn}
\setlength{\tabcolsep}{3pt}
\renewcommand{\arraystretch}{1.05}
\begin{tabular}{>{\centering\arraybackslash}p{0.95in}
                >{\raggedright\arraybackslash}p{2.18in}
                >{\centering\arraybackslash}p{0.45in}
                >{\centering\arraybackslash}p{1.20in}
                >{\centering\arraybackslash}p{0.85in}}
\toprule
\textbf{Level} & \textbf{Task} & \textbf{Samples} & \textbf{Source} & \textbf{Scoring} \\
\midrule
\multirow{8}{*}{\levelcell{L1}{Knowledge}} & \taskcell{Statute Recitation}{法条背诵} & 45 & LawBench (1-1) & Rule-based \\
 & \taskcell{Legal Rule Recall}{法条规则记忆} & 37 & In-house & Model-based \\
 & \taskcell{Legal Knowledge Error Detection}{法律知识幻觉} & 31 & In-house & Model-based \\
 & \taskcell{Applicable Law Identification}{适用法律推荐} & 35 & In-house & Model-based \\
 & \taskcell{Statutory Evolution Analysis}{立法沿革分析} & 57 & In-house & Model-based \\
 & \taskcell{Judicial Procedure QA}{司法程序问答} & 49 & In-house & Model-based \\
 & \taskcell{Legal Policy QA}{法律政策解答} & 10 & In-house & Model-based \\
 & \taskcell{Statute/regulation QA}{法律法规问答} & 42 & In-house & Model-based \\
\midrule
\multirow{11}{*}{\levelcell{L2}{Understanding}} & \taskcell{Legal Document Format Checking}{法律文书格式检查} & 13 & In-house & Model-based \\
 & \taskcell{Legal Grammar Checking}{法律专业语法检查} & 25 & In-house & Model-based \\
 & \taskcell{Case Event Timeline Extraction}{案情大事记梳理} & 25 & In-house & Model-based \\
 & \taskcell{Sensitive Information Redaction}{去除敏感信息} & 27 & In-house & Model-based \\
 & \taskcell{Judgment Reasoning Summarization}{判决书审理思路总结} & 20 & In-house & Model-based \\
 & \taskcell{Forensic Report Analysis}{司法专业报告分析} & 19 & In-house & Model-based \\
 & \taskcell{Legal Document Summarization}{法律文书摘要} & 20 & In-house & Model-based \\
 & \taskcell{Legal Literature Review}{法律文献综述} & 22 & In-house & Model-based \\
 & \taskcell{Legal Translation}{法律翻译} & 23 & In-house & Model-based \\
 & \taskcell{Legal Element Extraction}{法律要素还原} & 19 & In-house & Model-based \\
 & \taskcell{Long-Form Legal Document Understanding}{法律长文本分析总结} & 15 & In-house & Model-based \\
\midrule
\multirow{12}{*}{\levelcell{L3}{Analysis}} & \taskcell{Judicial Exam QA}{法律知识问答：司法考试题} & 44 & \sourcecell{Legal exams}{Judicial Exam, 24-25} & Model-based \\
 & \taskcell{Master-of-Law Exam QA}{法律知识问答：考研} & 10 & \sourcecell{Legal exams}{Master-of-Law Exam, 24-25} & Model-based \\
 & \taskcell{Legal Characterization}{法律定性判断} & 41 & In-house & Model-based \\
 & \taskcell{Fact-Finding Analysis}{事实认定分析} & 37 & In-house & Model-based \\
 & \taskcell{Disputed Issue Identification}{双方抗辩点识别} & 43 & In-house & Model-based \\
 & \taskcell{Copyright Infringement Text Comparison}{著作权侵权文本比对} & 43 & In-house & Model-based \\
 & \taskcell{Evidence Analysis}{证据分析} & 35 & In-house & Model-based \\
 & \taskcell{Trial-Focus Recommendation}{审理要点推荐} & 8 & In-house & Model-based \\
 & \taskcell{Legal Rationale Generation}{释法说理生成} & 15 & In-house & Model-based \\
 & \taskcell{Comparative Law Analysis}{比较法分析} & 19 & In-house & Model-based \\
 & \taskcell{Bid Qualification Extraction}{招投标文件资质提取} & 38 & In-house & Model-based \\
 & \taskcell{Commonsense Inconsistency Analysis}{违背常情常理分析} & 44 & In-house & Model-based \\
\bottomrule
\end{tabular}
\end{CJK*}
\caption{Task descriptions in \textsc{LegalRewardBench} (L1--L3). For different tasks, we use either rule-based or model-based automatic scoring methods based on the nature of the task.}
\label{tab:task_description}
\end{table*}

\begin{table*}[!h]
\centering
\small
\begin{CJK*}{UTF8}{gbsn}
\setlength{\tabcolsep}{3pt}
\renewcommand{\arraystretch}{1.05}
\begin{tabular}{>{\centering\arraybackslash}p{0.95in}
                >{\raggedright\arraybackslash}p{2.18in}
                >{\centering\arraybackslash}p{0.45in}
                >{\centering\arraybackslash}p{1.20in}
                >{\centering\arraybackslash}p{0.85in}}
\toprule
\textbf{Level} & \textbf{Task} & \textbf{Samples} & \textbf{Source} & \textbf{Scoring} \\
\midrule
\multirow{6}{*}{\levelcell{L4}{Decision}} & \taskcell{Sentence Prediction Without Statutes}{刑期预测：无法条内容} & 69 & LawBench (3-4) & Rule-based \\
 & \taskcell{Sentence Prediction With Statutes}{刑期预测：给定法条内容} & 68 & LawBench (3-5) & Rule-based \\
 & \taskcell{Judgment Outcome Prediction}{裁判结果预测} & 41 & In-house & Model-based \\
 & \taskcell{Public-Opinion Risk Assessment}{审判结果舆情预测} & 40 & In-house & Model-based \\
 & \taskcell{Expert Reasoning Outline Generation}{法律专家思路生成} & 36 & In-house & Model-based \\
 & \taskcell{Legal Task Planning}{法律任务规划} & 42 & In-house & Model-based \\
\midrule
\multirow{4}{*}{\levelcell{L5}{Writing}} & \taskcell{Procuratorial Recommendation Drafting}{检察建议书生成} & 11 & In-house & Model-based \\
 & \taskcell{Defense Pleading Drafting}{答辩状生成} & 43 & In-house & Model-based \\
 & \taskcell{Similar-Case Retrieval Report Drafting}{类案检索报告生成} & 28 & In-house & Model-based \\
 & \taskcell{Indictment Drafting}{起诉书生成} & 42 & In-house & Model-based \\
\bottomrule
\end{tabular}
\end{CJK*}
\caption{Task descriptions in \textsc{LegalRewardBench} (L4--L5, continued).}
\label{tab:task_description_cont}
\end{table*}

\section{Inference Details}
\label{app:inference_settings}

As explained in Section~\ref{sec:experiment-setting}, we evaluate three families of reward models under inference settings that match their model interfaces. \textbf{General GenRMs} are prompted as pairwise legal judges and are asked to output a verdict between two candidate responses, using the template shown in Prompt~\ref{prompt:generative_judge}. \textbf{Fine-tuned GenRMs} are evaluated with their native generative judgment templates whenever available, so that inference remains aligned with their supervised or preference-tuned training format. \textbf{Fine-tuned DisRMs} are used as scalar scorers: each query--response pair is scored independently, and the response with the higher score is selected as the model's pairwise preference.

For generative models, unless otherwise specified by the model provider, we use temperature 0.2, top-p 1.0, top-k disabled, a maximum generation length of 4096 new tokens, and a maximum context length of 32768 tokens. For Qwen3-series models, we follow the inference guideline with temperature 0.6, top-p 0.95, and top-k 20.

\section{Self-Bias of Reward Models}
\label{app:self_bias}


\begin{figure*}[!h]
\centering
\includegraphics[width=0.95\linewidth]{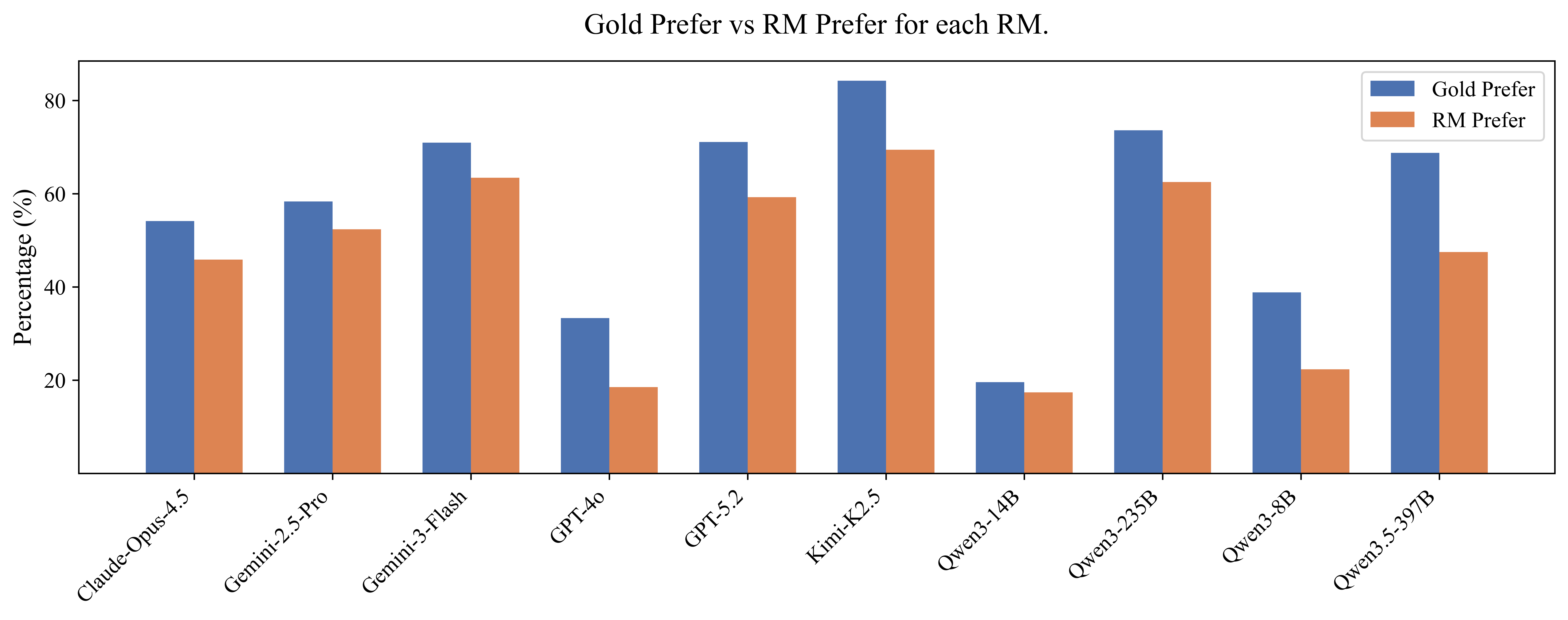}
\caption{Self-bias analysis for judge models. All values are reported as percentages.}
\label{tab:self_bias}
\end{figure*}

A common concern in RM evaluation is self-bias: if a model both generates candidate responses and acts as the RM, it may prefer its own style, reasoning, or wording \cite{wataoka2024self}. We therefore examine cases where one candidate response was generated by the RM itself. We record how often the RM selects its own response (\textit{RM Prefer}), how often the gold label prefers that response (\textit{Gold Prefer}), and define \textit{Self-bias} as their difference.

As shown in Figure~\ref{tab:self_bias}, all listed self-bias values are negative, indicating that these RMs do not systematically over-prefer their own outputs relative to the gold labels. This suggests that self-bias is not a dominant source of noise in the current \textsc{LegalRewardBench} setting.

\begin{figure*}[h]
\begin{promptbox}[title={Prompt~\thetcbcounter: Automatic Scoring Prompt},
label={prompt:auto_scoring}]
You are a professional judge for legal and related tasks. Please score the \texttt{candidate model response} according to the task requirement, rubrics, and the reference answer if provided. Use an integer scale from 0 to 100, where 100 is the full score.

\#\#\# Task Requirement / Query
\texttt{\{instruction\}}

\#\#\# Rubrics
\texttt{\{score\_point\}}

\texttt{\{trunc\_block\}}\texttt{\{ref\_block\}}

\#\#\# Candidate Model Response
\texttt{\{shown\}}

Please output your response in the following format:

\texttt{\{"reason": "Detailed reasoning process in Chinese explaining the reasons for deduction", "score": an integer score between 0 and 100\}}

Do not output any other text, explanation, or Markdown code-block markers.
\end{promptbox}
\end{figure*}

\begin{figure*}[h]
\begin{promptbox}[title={Prompt~\thetcbcounter: Training-Data Scoring Prompt},
label={prompt:training_data_scoring}]
You are a legal expert evaluator. Please score the model response according to the following information, using an integer score from 1 to 10:

\#\#\# Task Requirement
\texttt{\{instruction\}}

\#\#\# Query
\texttt{\{question\}}

\#\#\# Reference Answer
\texttt{\{reference\_answer\}}\texttt{\{analysis\_block\}}

\#\#\# Model Response
\texttt{\{model\_answer\}}

Scoring criteria:

- 1--3 points: The response is seriously incorrect or completely deviates from the task.

- 4--6 points: The response is partially correct but contains clear defects or omissions.

- 7--8 points: The response is mostly correct, with only minor flaws.

- 9--10 points: The response is fully correct, logically clear, and accurately expressed.

Please output only one JSON object:

\texttt{\{"score": <an integer from 1 to 10>, "reason": "<brief reason>"\}}
\end{promptbox}
\end{figure*}

\begin{figure*}[h]
\begin{promptbox}[title={Prompt~\thetcbcounter: Annotation Guideline for Short Outputs},
label={prompt:annotation_short}]
Please choose the better response to the query from the two candidate responses.

1. The task requires the model to first provide its reasoning process and then give the final response after \texttt{Final Output:}. Therefore, when annotating, please evaluate the response mainly based on the content after \texttt{Final Output:}, according to \texttt{score\_point}.

2. \texttt{score\_point} is the primary basis for comparison. However, if the two candidate responses are similar with respect to \texttt{score\_point} but one response has a clear quality problem, such as a serious professional error, you may annotate according to that quality issue.

3. Only when the final-output parts are close in quality should you compare the reasoning process before \texttt{Final Output:}.

4. If a response does not provide a reasoning process before the final response, it fails to follow the query instruction and should be penalized accordingly.

5. When the query is long, you may use a large language model to help analyze the query, but you must not ask a large language model to analyze the candidate responses.

6. If you think some rubrics in \texttt{score\_point} are problematic, please explain this in the notes.
\end{promptbox}
\end{figure*}

\begin{figure*}[h]
\begin{promptbox}[title={Prompt~\thetcbcounter: Annotation Guideline for Long Outputs},
label={prompt:annotation_long}]
Please choose the better response to the query from the two candidate responses.

1. The task requires the model to first provide its reasoning process and then give the final response after \texttt{Final Output:}. Therefore, when annotating, please evaluate the response mainly based on the content after \texttt{Final Output:}, according to \texttt{score\_point}.

2. Since there are many rubrics, we provide \texttt{score\_reason}, an external model's scoring rationale for each response with respect to \texttt{score\_point}. You may inspect \texttt{score\_reason} and use the external model's scoring process as a reference.

3. \texttt{score\_point} is the primary basis for comparison. However, if the two candidate responses are similar with respect to \texttt{score\_point} but one response has a clear quality problem, such as a serious professional error, you may annotate according to that quality issue.

4. Only when the final-output parts are close in quality should you compare the reasoning process before \texttt{Final Output:}.

5. If a response does not provide a reasoning process before the final response, it fails to follow the query instruction and should be penalized accordingly.

6. When the query is long, you may use a large language model to help analyze the query, but you must not ask a large language model to analyze the candidate responses.

7. If you think some rubrics in \texttt{score\_point} are problematic, please explain this in the notes.
\end{promptbox}
\end{figure*}

\begin{figure*}[h]
\begin{promptbox}[title={Prompt~\thetcbcounter: Generative Judge Inference Prompt},
label={prompt:generative_judge}]
Please act as an impartial legal expert judge and evaluate the quality of the responses provided by two AI assistants to the user's legal inquiry displayed below.

Begin your evaluation by comparing the two responses and provide a short explanation in Chinese.

Avoid any position biases and ensure that the order in which the responses were presented does not influence your decision.

Do not allow the length of the responses to influence your evaluation. Do not favor certain names of the assistants.

After providing your explanation, output your final verdict by strictly following this format: \texttt{[[A]]} if assistant A is better, \texttt{[[B]]} if assistant B is better.
\end{promptbox}
\end{figure*}

\section{Benchmark Examples}
\label{app:benchmark_examples}

To make the benchmark format concrete, we present three representative examples from \textsc{LegalRewardBench}: an L1 statute-recitation item from LawBench, an L3 judicial-exam item, and an L4 sentence-prediction item from LawBench. Each example follows the same structure: task instruction, query, rubrics, the chosen and rejected responses, and the reference answer. Note that due to space limitations, we omit the reasoning process of each candidate response. 

\newtcolorbox[auto counter, number within=section]{samplebox}[1][]{%
  enhanced,
  colback=white,
  colframe=gray!65!black,
  coltitle=black,
  colbacktitle=gray!15,
  fonttitle=\small\bfseries,
  toptitle=3pt,
  bottomtitle=3pt,
  boxrule=0.7pt,
  arc=1pt,
  left=3pt, right=3pt, top=3pt, bottom=3pt,
  title={Sample~\thetcbcounter},
  #1
}

\newcommand{\sampleblock}[3]{%
\vspace{2pt}
\begin{tcolorbox}[
  enhanced,
  colback=#2!4,
  colframe=#2!45!black,
  boxrule=0.35pt,
  arc=1pt,
  left=2pt, right=2pt, top=1.5pt, bottom=1.5pt,
  fontupper=\fontsize{5.6}{6.1}\selectfont,
  title={\textbf{#1}},
  colbacktitle=#2!14,
  coltitle=black,
  fonttitle=\fontsize{6}{6.4}\selectfont\bfseries,
  toptitle=1pt,
  bottomtitle=1pt
]
#3
\end{tcolorbox}}

\newcommand{\sampleanswer}[3]{%
\begin{tcolorbox}[
  enhanced,
  colback=#2!4,
  colframe=#2!45!black,
  boxrule=0.4pt,
  arc=1pt,
  left=2pt, right=2pt, top=1.5pt, bottom=1.5pt,
  fontupper=\fontsize{5.6}{6.1}\selectfont,
  title={\textbf{#1}},
  colbacktitle=#2!14,
  coltitle=black,
  fonttitle=\fontsize{6}{6.4}\selectfont\bfseries,
  toptitle=1pt,
  bottomtitle=1pt
]
#3
\end{tcolorbox}}

\newcommand{\sampleblockbi}[4]{%
\sampleblock{#1}{#2}{#3\par\medskip\textit{English:} #4}}

\newcommand{\sampleanswerbi}[4]{%
\sampleanswer{#1}{#2}{#3\par\medskip\textit{English:} #4}}

\begin{figure*}[!ht]
\begin{samplebox}[title={Sample~\thetcbcounter: L1 Knowledge -- Statute Recitation}, label={sample:benchmark_lawbench_l1}]
\begin{CJK*}{UTF8}{gbsn}
\sampleblockbi{Task Instruction}{gray}{回答以下问题，请先写出简要推理，换行后在新的一行以「最终输出：」开头直接给出法条内容。请你回答：}{Answer the following question. First provide a brief reasoning, and then, on a new line beginning with ``Final Output:'', directly provide the text of the relevant statutory provision.}
\sampleblockbi{Query}{gray}{社会法家庭教育促进法第二十二条的内容是什么？}{What does Article 22 of the Family Education Promotion Law provide?}
\sampleblockbi{Rubrics}{yellow}{最终输出应准确复述《中华人民共和国家庭教育促进法》第二十二条：未成年人的父母或者其他监护人应当合理安排未成年人学习、休息、娱乐和体育锻炼的时间，避免加重未成年人学习负担，预防未成年人沉迷网络。}{The final output should accurately reproduce Article 22 of the Family Education Promotion Law of the People's Republic of China: Parents or other guardians of minors shall reasonably arrange the minors' time for study, rest, recreation, and physical exercise, avoid increasing their study burden, and prevent minors from becoming addicted to the Internet.}
\begin{tcbraster}[raster columns=2, raster equal height=rows, raster column skip=5pt]
\sampleanswerbi{Chosen Response}{green}{最终输出：未成年人的父母或者其他监护人应当合理安排未成年人学习、休息、娱乐和体育锻炼的时间，避免加重未成年人学习负担，预防未成年人沉迷网络。}{Final Output: Parents or other guardians of minors shall reasonably arrange the minors' time for study, rest, recreation, and physical exercise, avoid increasing their study burden, and prevent minors from becoming addicted to the Internet.}
\sampleanswerbi{Rejected Response}{red}{最终输出：父母或者其他监护人应当树立家庭是第一个课堂、家长是第一任老师的责任意识，承担对未成年子女实施家庭教育的主体责任。父母或者其他监护人应当亲自养育未成年子女，加强亲子陪伴。父母或者其他监护人应当共同参与家庭教育，关注未成年子女的生理、心理、智力发展状况，尊重其参与相关家庭事务和发表意见的权利。父母或者其他监护人不得实施家庭暴力，不得因性别、身体状况、智力状况等歧视未成年子女，不得胁迫、引诱、教唆、纵容、包庇未成年人实施违法犯罪行为，不得放任、教唆或者帮助未成年人自杀、自残、欺凌、暴力等行为。}{Final Output: Parents or other guardians shall recognize that the family is the first classroom and that parents are the first teachers, and shall assume primary responsibility for providing family education to their minor children. Parents or other guardians shall personally raise their minor children and strengthen parent-child companionship. Parents or other guardians shall jointly participate in family education, pay attention to the physical, psychological, and intellectual development of their minor children, and respect their right to participate in relevant family matters and express their opinions. Parents or other guardians shall not commit domestic violence or discriminate against their minor children on the basis of sex, physical condition, intellectual condition, or other factors; coerce, entice, instigate, tolerate, or shelter minors in committing unlawful or criminal acts; or allow, instigate, or assist minors in engaging in suicide, self-harm, bullying, violence, or other such acts.}
\end{tcbraster}
\sampleblockbi{Reference Answer}{blue}{未成年人的父母或者其他监护人应当合理安排未成年人学习、休息、娱乐和体育锻炼的时间，避免加重未成年人学习负担，预防未成年人沉迷网络。}{Parents or other guardians of minors shall reasonably arrange the minors' time for study, rest, recreation, and physical exercise, avoid increasing their study burden, and prevent minors from becoming addicted to the Internet.}
\end{CJK*}
\end{samplebox}
\end{figure*}

\begin{figure*}[!ht]
\begin{samplebox}[title={Sample~\thetcbcounter: L3 Analysis -- Judicial Exam QA}, label={sample:benchmark_exam_l3}]
\begin{CJK*}{UTF8}{gbsn}
\sampleblockbi{Task Instruction}{gray}{题干与具体问题见用户消息。请先进行分析或推理，然后在新的一行「最终输出：」开头给出你的作答。}{The case facts and specific question are provided below. First analyze or reason through the problem, and then provide your answer on a new line beginning with ``Final Output:''.}
\sampleblockbi{Query}{gray}{【案情】某应急管理局因为某公司的焊接工人没有取得焊接证，责令其临时停业，取得证件后再进行生产。后逾期没有取得许可证，管理局拟作出行政处罚决定，责令停业，罚款7万元。告知其5日内可以进行申诉等等。但是该公司5日内没有申诉，管理局就做出行政处罚。后该公司觉得处罚不合理，申请复议。复议机关作出维持决定。\par 该公司如果逾期不缴纳罚款，丙区应急管理局应该如何处理？}{Case facts: An emergency management bureau ordered a company to temporarily suspend operations because its welding workers had not obtained the required welding certificates, and allowed it to resume production only after the certificates were obtained. The company subsequently failed to obtain the required permit within the prescribed period, and the bureau proposed an administrative penalty requiring it to suspend operations and pay a fine of 70,000 yuan. The company was notified that it could file an appeal within five days, but did not do so within that period, and the bureau subsequently issued the administrative penalty. The company later considered the penalty unreasonable and applied for administrative reconsideration. The reconsideration authority upheld the decision.\par If the company fails to pay the fine within the prescribed period, how should the emergency management bureau of District C handle the matter?}
\sampleblockbi{Rubrics}{yellow}{1. 说明应当书面催告，并载明金额、给付方式和期限；\par 2. 说明逾期不缴纳罚款的，应按罚款数额每日百分之三加处罚款，并告知该标准，且加处罚款的数额不得超过罚款数额；\par 3. 说明若在法定期限内不申请行政复议或者提起行政诉讼，又不履行行政决定的，因该局无权直接执行罚款，应自期限届满之日起三个月内向本机关所在地法院申请强制执行。}{1. State that the bureau should issue a written demand specifying the amount, the method of payment, and the time limit.\par 2. State that if the fine is not paid on time, an additional fine of three percent of the original fine amount per day shall be imposed, that this standard should be communicated to the company, and that the additional fine may not exceed the original fine amount.\par 3. State that if the company neither applies for administrative reconsideration nor brings an administrative lawsuit within the statutory period, and also fails to comply with the administrative decision, the bureau, which has no authority to directly enforce the fine, should apply to the court at the bureau's location for compulsory enforcement within three months after the statutory period expires.}
\begin{tcbraster}[raster columns=2, raster equal height=rows, raster column skip=5pt]
\sampleanswerbi{Chosen Response}{green}{最终输出：丙区应急管理局可以每日按罚款数额的3\%加处罚款（加处罚款额不得超过7万元罚款本数）；在申请强制执行前应当依法书面催告；若该公司在法定期限内不提起诉讼又不履行，应急管理局应当申请人民法院强制执行。}{Final Output: The emergency management bureau of District C may impose an additional fine at a rate of 3\% of the original fine amount per day, provided that the additional fine does not exceed the original fine of 70,000 yuan. Before applying for compulsory enforcement, it should issue a written demand in accordance with the law. If the company neither brings a lawsuit within the statutory period nor complies with the decision, the bureau should apply to a people's court for compulsory enforcement.}
\sampleanswerbi{Rejected Response}{red}{最终输出：丙区应急管理局应当按下列步骤处理：\par 1. 催告缴纳：向该公司送达缴纳罚款的催告或通知，明确补缴期限、缴纳方式及逾期后果。\par 2. 加处罚款：公司逾期仍不缴纳的，可以依法对逾期未缴部分按日加处罚款，通常为按日加收不超过罚款数额一定比例的加处罚款。\par 3. 申请法院强制执行：经催告后仍不履行的，应当向有管辖权的人民法院申请强制执行，而不是自行强制扣划或查封，除非有明确法律授权。}{Final Output: The emergency management bureau of District C should proceed as follows.\par 1. Demand payment: Serve the company with a written demand or notice to pay the fine, specifying the time limit for payment, the method of payment, and the consequences of late payment.\par 2. Impose an additional fine: If the company still fails to pay on time, an additional daily fine may be imposed on the unpaid amount in accordance with the law, generally at a rate not exceeding a specified proportion of the original fine.\par 3. Apply for compulsory enforcement: If the company still fails to comply after receiving the demand, the bureau should apply to the competent people's court for compulsory enforcement rather than directly freezing or seizing the company's assets, unless expressly authorized to do so by law.}
\end{tcbraster}
\sampleblockbi{Reference Answer}{blue}{1. 书面催告，载明金额、给付方式和期限。\par 2. 该公司如果逾期不缴纳罚款的，该区应急管理局应按罚款数额的每日百分之三加处罚款，告知标准且加处罚款的数额不得超过罚款的数额。\par 3. 若该公司在法定期限内不申请行政复议或者提起行政诉讼，又不履行行政决定的，因为该区应急管理局无权直接执行罚款，其可以自期限届满之日起三个月内，向本机关所在地法院申请强制执行。}{1. Issue a written demand specifying the amount, the method of payment, and the time limit.\par 2. If the company fails to pay the fine on time, the district emergency management bureau shall impose an additional fine at a rate of three percent of the fine amount per day, notify the company of this standard, and ensure that the additional fine does not exceed the original fine amount.\par 3. If the company neither applies for administrative reconsideration nor brings an administrative lawsuit within the statutory period, and also fails to comply with the administrative decision, the bureau, which has no authority to directly enforce the fine, may apply to the court at the bureau's location for compulsory enforcement within three months after the statutory period expires.}
\end{CJK*}
\end{samplebox}
\end{figure*}

\begin{figure*}[!ht]
\begin{samplebox}[title={Sample~\thetcbcounter: L4 Decision -- Sentence Prediction Without Statutes}, label={sample:benchmark_lawbench_l4}]
\begin{CJK*}{UTF8}{gbsn}
\sampleblockbi{Task Instruction}{gray}{根据下列事实、罪名和刑法法条预测判决刑期。请先进行分析或推理，然后给出判决刑期为多少月，请将答案填在[刑期]与\textless eoa\textgreater 之间。例如[刑期]12月\textless eoa\textgreater 。请你回答：}{Based on the following facts, charge, and provision of the Criminal Law, predict the length of the sentence in months. First analyze or reason through the case, and then place the predicted sentence term between [刑期] and \textless eoa\textgreater. For example, [刑期]12月\textless eoa\textgreater.}
\sampleblockbi{Query}{gray}{事实：新沂市人民检察院指控：2010年4月至2015年间，被告人潘某某在新沂市＊＊镇＊＊村＊组＊＊号其家中开设诊所，因未取得《医疗机构执业许可证》及《执业医师资格证》行医，分别于2010年6月1日、2011年5月27日、2012年5月20日被新沂市卫生局行政处罚。后被告人潘某某继续在其诊所内无证行医，并于2015年10月30日正在行医时被新沂市公安局马陵山派出所查获。被告人潘某某于2015年10月30日被新沂市公安局马陵山派出所民警依法传唤到案，其如实供述上述犯罪事实。罪名：非法行医。法条：刑法第336条。}{Facts: The Xinyi Municipal People's Procuratorate charged that, from April 2010 to 2015, defendant Pan operated a clinic at his home in Xinyi without obtaining a Medical Institution Practice License or a Physician Qualification Certificate. The Xinyi Municipal Health Bureau imposed administrative penalties on him on June 1, 2010, May 27, 2011, and May 20, 2012. Pan subsequently continued to practice medicine without a license at his clinic and was apprehended while practicing medicine on October 30, 2015, by officers from the Malingshan Police Station of the Xinyi Public Security Bureau. On the same day, Pan was lawfully summoned to appear before officers from the Malingshan Police Station and truthfully confessed to the above criminal conduct. Charge: illegal practice of medicine. Provision: Article 336 of the Criminal Law.}
\sampleblockbi{Rubrics}{yellow}{参考刑期为16个月。从100分起，预测刑期与参考刑期每相差1个月扣1分，分数下限为0。}{The reference sentence is 16 months. Starting from a score of 100, one point is deducted for each month of absolute difference between the predicted sentence and the reference sentence, with a minimum score of 0.}
\begin{tcbraster}[raster columns=2, raster equal height=rows, raster column skip=5pt]
\sampleanswerbi{Chosen Response}{green}{[刑期]18月\textless eoa\textgreater}{[刑期]18月\textless eoa\textgreater}
\sampleanswerbi{Rejected Response}{red}{[刑期]30月\textless eoa\textgreater}{[刑期]30月\textless eoa\textgreater}
\end{tcbraster}
\sampleblockbi{Reference Answer}{blue}{刑期：16个月。}{Sentence term: 16 months.}
\end{CJK*}
\end{samplebox}
\end{figure*}

\label{sec:appendix}
\end{document}